\documentclass{article}

\PassOptionsToPackage{numbers,sort&compress}{natbib}

\usepackage[main,final]{neurips_2026}

\usepackage[utf8]{inputenc}
\usepackage[T1]{fontenc}
\usepackage{hyperref}
\usepackage{url}
\usepackage{booktabs}
\usepackage{amsfonts}
\usepackage{amsmath}
\usepackage{amssymb}
\usepackage{nicefrac}
\usepackage{microtype}
\usepackage{graphicx}
\usepackage{multirow}
\usepackage{array}
\usepackage{pifont}
\usepackage{subcaption}
\usepackage[table]{xcolor}

\hypersetup{
  colorlinks=true,
  urlcolor=blue
}

\definecolor{diagrow}{RGB}{238,244,255}
\definecolor{oursrow}{RGB}{232,245,233}

\newcommand{\dataset}{\textsc{CineScript}}
\newcommand{\ours}{\textsc{CineGEN}}

\newcommand{\dirspeed}{\textsc{DirSpeed}}
\newcommand{\pose}{\textsc{Pose9D}}
\newcommand{\dsmath}{\mathrm{ds}}
\newcommand{\posemath}{\mathrm{pose}}

\newcommand{\alnscore}{AlignScore}
\newcommand{\fcd}{FCD}
\newcommand{\director}{\textsc{E.T.}}
\newcommand{\gendop}{\textsc{GenDoP}}
\newcommand{\ccd}{\textsc{CCD}}
\newcommand{\up}{\,$\uparrow$}
\newcommand{\down}{\,$\downarrow$}
\newcommand{\cmark}{\ding{51}}
\newcommand{\xmark}{\ding{55}}

\newcommand{\pretrainedmark}{\ensuremath{^{*}}}
\newcommand{\dirspeedmark}{\ensuremath{^{\dagger}}}

\newcommand{\NumHumanShots}{30}
\newcommand{\NumHumanMethods}{7}
\newcommand{\NumHumanParticipants}{24}
\newcommand{\NumHumanJudgements}{720}

\title{Unveiling the Value of Motion for \\ Cinematic Camera Trajectories}

\author{%
  \begin{tabular}{ccc}
    \textbf{Ziqi Zhou}\textsuperscript{1} &
    \textbf{Yujian Yuan}\textsuperscript{2} &
    \textbf{Laura Sevilla-Lara}\textsuperscript{1} \\
    \multicolumn{3}{c}{\small\normalfont
      \textsuperscript{1}University of Edinburgh \quad
      \textsuperscript{2}The Hong Kong University of Science and Technology} \\
    \small\normalfont\texttt{Ziqi.Zhou@ed.ac.uk} &
    \small\normalfont\texttt{yyuanbn@connect.ust.hk} &
    \small\normalfont\texttt{l.sevilla@ed.ac.uk}
  \end{tabular}
}

\begin{document}

\maketitle

\begin{abstract}
Cinematic camera motion is a fundamental storytelling tool, defined not only by \emph{where} the camera is positioned in the scene, but also by \emph{how} it moves in terms of direction and speed. Recent work on camera trajectory generation and alignment to text relies on pose-centric representations. While in principle a network could derive direction of movement and speed, we find that in practice this might not happen. In fact, in this paper we discover that decomposing the camera trajectory representation from the traditional per-frame poses to direction and speed has surprising benefits across multiple tasks, including trajectory-to-text alignment as well as text-to-trajectory generation. To accurately evaluate the former, we introduce a simple and reliable protocol that overcomes the limitations of prior evaluation baselines. For the latter, building on this representational insight, we propose a novel generative model for camera trajectories, \ours, that achieves superior performance across a variety of metrics. We also propose a novel dataset, \dataset, containing movie clips that are enriched with scene descriptions as well as higher-level metadata. This novel data allows us to test models' ability to capture high-level cinematographic information. We show that, despite its simplicity, representing camera trajectories through direction and speed not only helps numerically to achieve better alignment and generation, but also inherently encodes complex directorial intent.
Feel free to visit our \href{https://jia1018.github.io/CineGEN/}{project page} for more information.
\end{abstract}

\section{Introduction}
\label{sec:intro}

Camera motion plays a vital role in cinematic storytelling, shaping visual pacing, audience attention, and directorial style. Driven by applications in virtual production~\cite{christie2015toric, galvane2018directing} and controllable video generation~\cite{bai2025recammaster, yu2025trajectorycrafter, wang2024motionctrl, he2024cameractrl, guo2024animatediff}, recent text-conditioned camera trajectory generators have made significant progress in synthesizing camera paths from natural-language descriptions~\cite{jiang2024ccd,courant2024et,zhang2025gendop,courant2025pulpmotion}. However, despite these advances, a fundamental challenge remains: how can we represent camera trajectories in a way that naturally aligns with human language?

Most existing methods parameterize camera trajectories as sequences of absolute per-frame poses~\cite{jiang2024ccd,courant2024et,zhang2025gendop, courant2025pulpmotion}. While geometrically complete, this pose-centric approach fundamentally clashes with the nature of cinematographic language. Human descriptions emphasize \emph{how} a camera moves, such as ``gradually dollies in and pans right'', rather than \emph{where} it is positioned in a global 3D space. By forcing semantic motion instructions into an absolute coordinate system, current models unnecessarily entangle direction, speed, and spatial location into a single opaque vector. This representational bottleneck severely complicates both text-to-trajectory generation and cross-modal alignment.

We argue that trajectory representation is not a minor implementation detail, but the central key to unlocking controllable cinematic generation. To overcome the geometric entanglement, we propose a remarkably simple yet highly effective shift to a motion-centric perspective. We introduce \dirspeed, a parameterization that converts absolute poses into frame-to-frame velocities, explicitly decoupling them into normalized direction and log-speed. By mathematically isolating the axis of movement from its magnitude, this decomposition naturally mirrors the human vocabulary of cinematography. By natively aligning the physical action space with human descriptions, \dirspeed\ frees the network from implicitly disentangling global coordinates, transforming a complex reasoning problem into a direct geometric mapping.

Moving beyond theoretical representation, evaluating a model's grasp of real-world cinematography demands a foundation of deeply contextualized data. We therefore construct \dataset, a comprehensive real-movie benchmark comprising approximately 28K clips and 10M frames drawn from diverse public sources~\cite{liu2025shotbench,wang2025cinetechbench,rao2020movieshots,bain2020condensedmovies,qiao2025vadb}. Going beyond prior datasets that rely solely on local motion captions~\cite{jiang2024ccd,courant2024et,zhang2025gendop}, \dataset\ provides screenplay-style loglines, which act as concise textual summaries of scene context and narrative action, as well as real movie-level global attributes (e.g., era, genre, director) retrieved by linking identifiable clips to public knowledge bases. This rich annotation enables the study of scene-aware generation and opens up the exploration of higher-level cinematic attributes---an indispensable yet historically overlooked dimension of camera motion.

While \dataset\ unlocks the exploration of high-level cinematic concepts, a critical bottleneck remains in evaluating the foundational alignment between camera trajectories and local motion descriptions. Reliably quantifying this core motion--text alignment calls for a dedicated discriminative protocol. However, prior baselines (e.g., CLaTr~\cite{courant2024et, zhang2025gendop, courant2025pulpmotion, petrovich23tmr}) often entangle text alignment with heavy trajectory-reconstruction objectives, which can obscure true instance-level correspondence. By stripping away this generative overhead and establishing a lightweight, purely contrastive setup, we explicitly unveil the value of the motion-centric approach, demonstrating that \dirspeed\ dramatically improves text--trajectory retrieval over pose-based alternatives regardless of the underlying evaluator. Building on this validated representation, we propose \ours, a novel generative model for camera trajectories. Existing standard diffusion models~\cite{jiang2024ccd,courant2024et} lack a mechanism for progressive motion commitment, while strictly causal autoregressive models~\cite{zhang2025gendop} rely on a unidirectional generation order that may limit the flexibility of global cinematic planning. To bridge this gap, \ours\ employs a text-conditioned masked autoregressive (MAR) architecture ~\cite{chang2022maskgit,li2024mar,fan2025fluid}. This formulation provides the best of both worlds, naturally balancing progressive step-by-step commitment with the bidirectional context necessary to satisfy global cinematic constraints.

Our experiments  unveil the hidden fundamental value of motion-centric representations in this task. Under \dirspeed, text alignment improves noticeably compared to pose-based alternatives, and our contrastive evaluation protocol establishes a strictly more reliable evaluation space than CLaTr. Building on these insights, \ours\ achieves superior performance in both trajectory quality and text alignment. Furthermore, \dataset\ enables us to step beyond local motion descriptions and explore the previously overlooked correlation between camera motion and high-level movie attributes. Our probes confirm that the generated trajectories successfully preserve broader cinematic patterns, such as era, genre, and directorial style, highlighting exciting new avenues for future exploration. Ultimately, these integrated components establish a thoroughly language-aligned framework for understanding and generating cinematic camera motion.

\section{Related Work}
\label{sec:related}

\noindent\textbf{Camera trajectory generation.}
Recent research has shifted from rule-based virtual cinematography to data-driven generative models. \ccd~\cite{jiang2024ccd} pioneered text-conditioned trajectory diffusion, which E.T.~\cite{courant2024et} later extended to real-movie data alongside the CLaTr evaluation metric. GenDoP~\cite{zhang2025gendop} further advanced this domain using causal autoregressive models trained on real-world datasets. Concurrently, the scope of trajectory generation has expanded significantly: Director3D~\cite{li2024director3d} jointly generates trajectories and 3D scenes, PulpMotion~\cite{courant2025pulpmotion} enforces actor-camera coherence, and ShotVerse~\cite{yang2026shotverse} tackles multi-shot cinematic planning. In adjacent embodied domains, NWM~\cite{bar2024navigation} and DVGFormer~\cite{hou2024dvgformer} explore navigation and drone control. Additionally, studies like Seeing without Pixels~\cite{xue2025seeing} highlight that trajectories inherently possess semantic meaning alignable with language. While these diverse efforts primarily focus on novel generative architectures or expanded task definitions, they largely inherit pose-centric parameterizations. Our work departs from this norm by identifying the underlying trajectory representation as a critical bottleneck, demonstrating that a motion-centric reformulation can fundamentally bridge the semantic gap between physical geometry and natural language.

\noindent\textbf{Masked autoregressive models.}
Masked autoregressive (MAR) models combine the bidirectional context of masked modeling with the progressive commitment of autoregressive decoding. Initially popularized for discrete visual tokens through works like MaskGIT~\cite{chang2022maskgit}, Muse~\cite{chang2023muse}, and MAGVIT~\cite{yu2023magvit}, the paradigm has recently shifted toward continuous-token generation. Methods such as GIVT~\cite{tschannen2024givt}, MAR~\cite{li2024mar}, and Fluid~\cite{fan2025fluid} bypass vector quantization entirely, directly modeling real-valued sequences with diffusion-based prediction heads. This continuous approach has demonstrated strong scalability for complex spatiotemporal and planning tasks~\cite{yao2025denoising,liu2024mardini,zhou2025magi}. Our approach extends this continuous MAR paradigm to the domain of camera trajectory generation. However, unlike image and video models that must rely on lossy autoencoders to compress high-dimensional visual vocabularies, cinematic camera trajectories are natively compact. We exploit this unique property to apply masked autoregression directly in the continuous feature space, enabling a generation process that naturally balances progressive motion commitment with bidirectional cinematic context.

\begin{figure}[t]
  \centering
  \includegraphics[width=\linewidth]{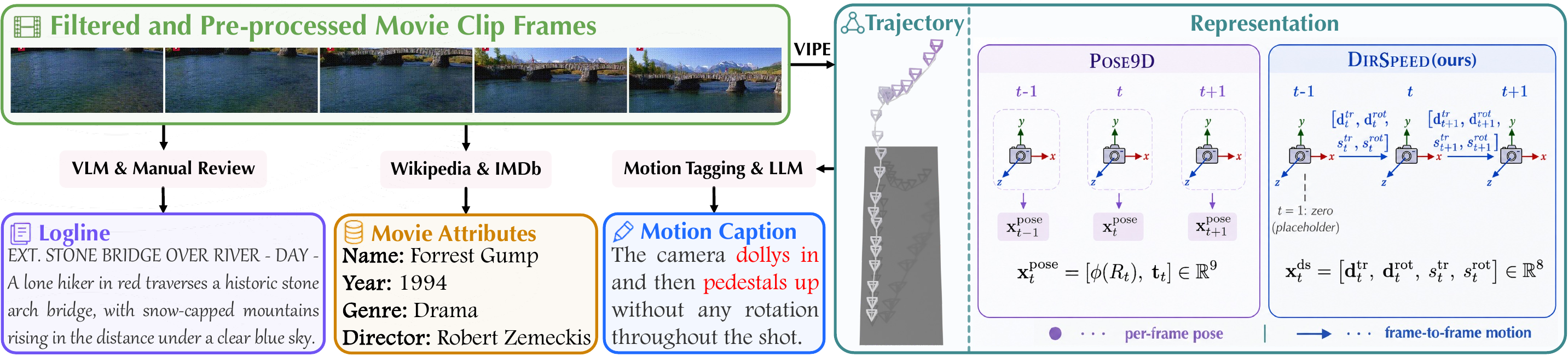}
  \caption{Overview of \dataset\ construction and trajectory representation. Movie clips are re-processed into camera trajectories, motion captions, screenplay-style loglines, and linked movie attributes. The right panel illustrates two kinds of representations mentioned in our paper. 
  }
  \label{fig:data_construction}
\end{figure}

\section{\dirspeed: A Motion-Centric Trajectory Representation}
\label{sec:repr}

As established in Sec.~\ref{sec:intro}, the choice of trajectory representation is not a minor implementation detail; it fundamentally dictates how well a model can align visual motion with natural language. Most prior work parameterizes camera trajectories using absolute per-frame poses. We compare this standard pose representation, denoted by \pose, with our proposed motion-centric direction-speed representation, denoted by \dirspeed. Fig.~\ref{fig:data_construction} provides a visual demonstration of the two approaches.

\noindent\textbf{Original pose parameterization (\pose).}
For each frame $t$, let the camera pose be given by a rotation matrix $R_t\in SO(3)$ and a translation vector $\mathbf{t}_t\in\mathbb{R}^3$. Following prior work, we represent the rotation by its continuous $6$D form~\cite{zhou2019rot6d}, denoted by $\phi(R_t)\in\mathbb{R}^6$, and concatenate it with translation as $\mathbf{x}^{\posemath}_{t} = \left[ \phi(R_t),\; \mathbf{t}_t \right] \in \mathbb{R}^9$. 
While this formulation is geometrically complete, it forces models to implicitly deduce relative motion from absolute coordinates, leading to the geometric entanglement that makes text--trajectory alignment difficult.

\noindent\textbf{Direction-speed parameterization (\dirspeed).}
To bridge the gap between physical geometry and natural language, we introduce \dirspeed, a representation designed to be both semantically aligned and numerically robust. We begin with the intuition that human motion captions describe frame-to-frame camera behavior---specifically, the direction and speed of movement---rather than absolute spatial coordinates. Driven by this semantic alignment, we first compute the translational and rotational velocities from the pose sequence $\{(R_t,\mathbf{t}_t)\}_{t=1}^T$ for $t=2,\ldots,T$:
$$ \Delta \mathbf{t}_t = \mathbf{t}_t-\mathbf{t}_{t-1}, \qquad \boldsymbol{\omega}_t = \log_{SO(3)}(R_{t-1}^{\top}R_t), $$
where $\log_{SO(3)}$ maps a relative rotation matrix to its axis-angle vector. To align sequence lengths, we set $\Delta\mathbf{t}_1=\mathbf{0}$ and $\boldsymbol{\omega}_1=\mathbf{0}$ as zero placeholders.

Crucially, raw velocity vectors still entangle the direction of movement with its overall speed. To isolate the exact motion factors described by language, we explicitly decompose each velocity vector into a normalized unit direction and a scalar speed:
$$ \mathbf{d}^{\mathrm{tr}}_t = \frac{\Delta\mathbf{t}_t}{\|\Delta\mathbf{t}_t\|+\varepsilon}, \qquad s^{\mathrm{tr}}_t = \log(\|\Delta\mathbf{t}_t\|+\varepsilon), $$
$$ \mathbf{d}^{\mathrm{rot}}_t = \frac{\boldsymbol{\omega}_t}{\|\boldsymbol{\omega}_t\|+\varepsilon}, \qquad s^{\mathrm{rot}}_t = \log(\|\boldsymbol{\omega}_t\|+\varepsilon). $$
Here, $\mathbf{d}^{\mathrm{tr}}_t,\mathbf{d}^{\mathrm{rot}}_t\in\mathbb{R}^3$ encode the pure directions of translation and rotation, with normalization explicitly isolating geometric orientation from scale. For the speed components $s^{\mathrm{tr}}_t$ and $s^{\mathrm{rot}}_t$, we apply a logarithmic transformation to compress the massive dynamic range of real cinematic movements into a stable distribution. The final per-step feature vector concatenates these components as $\mathbf{x}^{\dsmath}_t = \left[ \mathbf{d}^{\mathrm{tr}}_t,\; \mathbf{d}^{\mathrm{rot}}_t,\; s^{\mathrm{tr}}_t,\; s^{\mathrm{rot}}_t \right] \in \mathbb{R}^8$. Ultimately, this explicit decoupling of bounded directions and log-compressed speeds provides a natively language-aligned and numerically stable foundation that significantly eases neural network optimization, which we empirically validate in Sec.~\ref{sec:experiments}.

\section{The \dataset\ Dataset}
\label{sec:dataset}


Existing datasets~\cite{jiang2024ccd, courant2024et, zhang2025gendop, courant2025pulpmotion} for text-conditioned camera trajectory generation mainly provide motion-level supervision, pairing trajectories solely with direct movement descriptions. To ground our study in actual filmmaking and support scene-aware exploration, we construct \dataset, which extends this setting with screenplay-style loglines and linked real-movie attributes.


Fig.~\ref{fig:data_construction} illustrates the construction of \dataset. Starting from heterogeneous public movie-clip datasets, we first apply quality filtering and spatial pre-processing to ensure visual consistency (detailed in the following filtering step). We then extract per-frame camera trajectories using VIPE~\cite{vipe}, generate motion captions via motion tagging and large language models (LLMs), produce screenplay-style loglines using vision-language models (VLMs), and link identifiable clips to real movie metadata via public knowledge bases (e.g., Wikipedia and IMDb). This pipeline ensures that each retained clip is associated with synchronized visual content, camera motion, motion-level text, scene-level context, and, when available, higher-level movie attributes.

\noindent\emph{Clip collection and filtering.}
To ensure a diverse distribution of cinematic styles, shot patterns, and scene content, we construct \dataset\ from movie clips drawn from five public datasets: ShotBench~\cite{liu2025shotbench}, CineTechBench~\cite{wang2025cinetechbench}, MovieShots~\cite{rao2020movieshots}, CMD~\cite{bain2020condensedmovies}, and the film-derived subset of VADB~\cite{qiao2025vadb}. Because these sources are heterogeneous in scale and provenance, we re-process all clips through a unified pipeline rather than using any source annotations as-is. During this stage, we apply rigorous quality-control filters to remove clips with advertisement or UI artifacts, discard low-quality videos, and exclude abnormal aspect ratios. We also crop black borders prior to pose extraction to cleanly isolate the core content regions.

\noindent\emph{Camera trajectories.}
For each retained clip, we extract per-frame camera poses using VIPE~\cite{vipe}. The extracted trajectories are subsequently cleaned, smoothed, and converted into fixed-length sequences by truncating clips above the 90th length percentile and padding shorter clips with masks.

\noindent\emph{Motion captions.}
We generate camera-motion captions following the motion-tagging and caption-generation procedure of E.T.~\cite{courant2024et}. Specifically, each trajectory is segmented into temporally coherent motion primitives using velocity-based thresholding. The resulting structured tags are converted into natural-language descriptions by an LLM (Mistral-7B~\cite{jiang2023mistral}), such as \emph{``a steady dolly-in with a pan right.''} This annotation captures exactly what the camera does, serving as the primary supervision for motion-level controllability.

\noindent\emph{Loglines.}
Crucially, each clip is also paired with a screenplay-style logline describing the visual scene. Generated using Qwen3-VL-32B-Instruct~\cite{qwen3vl} via a structured prompt and manually reviewed for consistency, the final loglines follow the format \emph{[INT./EXT.] [Location] -- [Time] -- [Action]} (see Fig.~\ref{fig:data_construction} for an example). Unlike motion captions, loglines summarize spatial and narrative context without describing the camera movement itself. Because coarse motion captions alone cannot capture the full physical nuance of a camera path, this complementary text allows us to explore how the exact execution of a given motion naturally adapts to different scene contexts, revealing stylistic subtleties that motion text inherently misses. Details for logline generation are provided in Appendix~\ref{app:logline_generation}.

\noindent\emph{Movie attribute linking.}
Finally, for clips whose filenames preserve identifiable information (e.g., movie title or IMDb ID), we link them to real movie metadata through public databases (Wikidata, Wikipedia, and IMDb). We retrieve  attributes such as release year, genre, and director. This metadata allows us to go beyond local motion evaluation and explore the often-overlooked correlation between camera behavior and macro-level cinematic style.


\paragraph{Dataset Statistics.}
\dataset\ comprises approximately $28$K clips and $10$M frames, with an average duration of $12.0$ seconds per clip. Compared to prior datasets~\cite{jiang2024ccd, courant2024et, zhang2025gendop, courant2025pulpmotion}, \dataset\ is the first one to jointly contain motion captions, scene loglines, and real movie attributes (in a subset of the clips). This resulting metadata-linked subset contains $3{,}163$ clips from roughly $1{,}400$ unique films.

\begin{figure}[t]
  \centering
  \begin{subfigure}{0.58\linewidth}
    \centering
    \includegraphics[width=\linewidth]{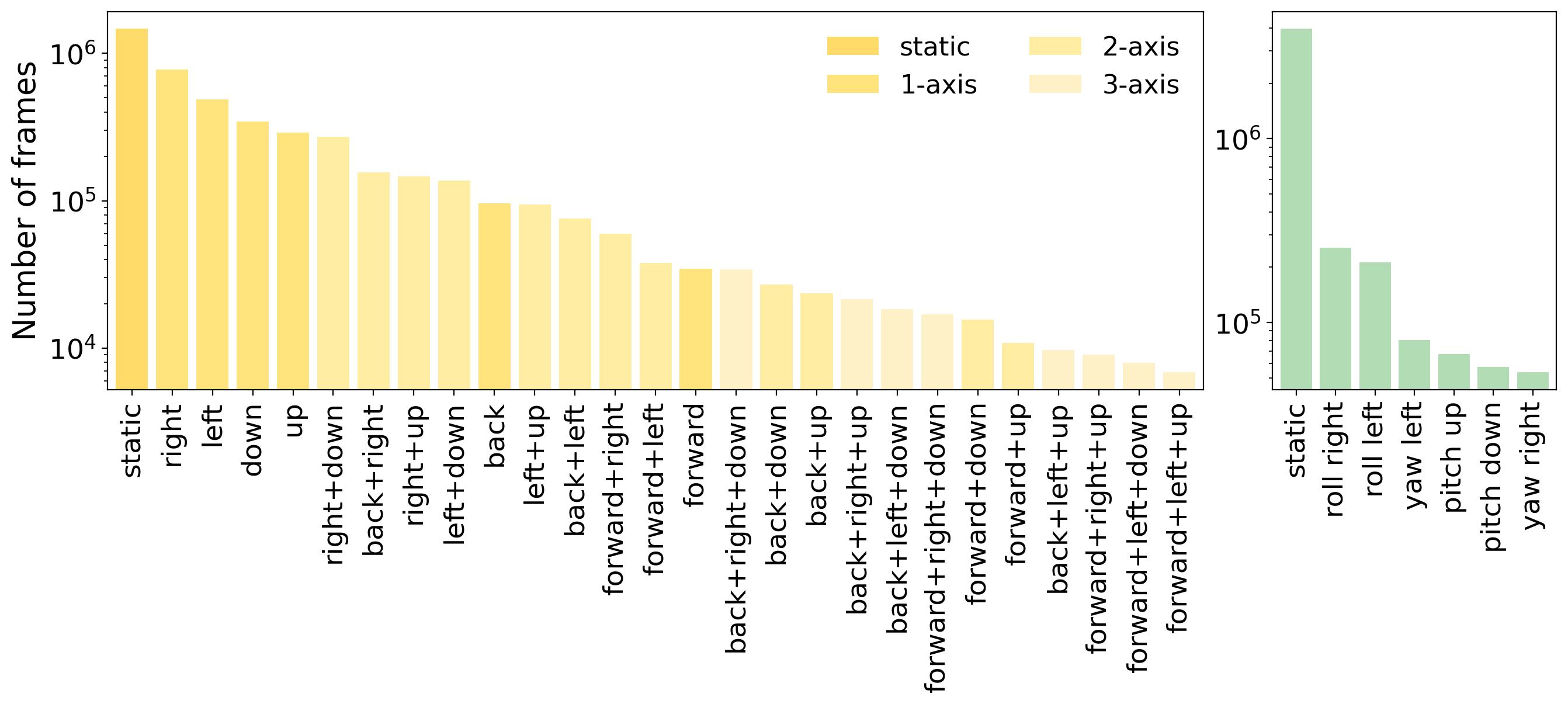}
    \caption{Motion primitive distribution}
    \label{fig:stat_motion_primitives}
  \end{subfigure}
  \hfill
  \begin{subfigure}{0.41\linewidth}
    \centering
    \includegraphics[width=\linewidth]{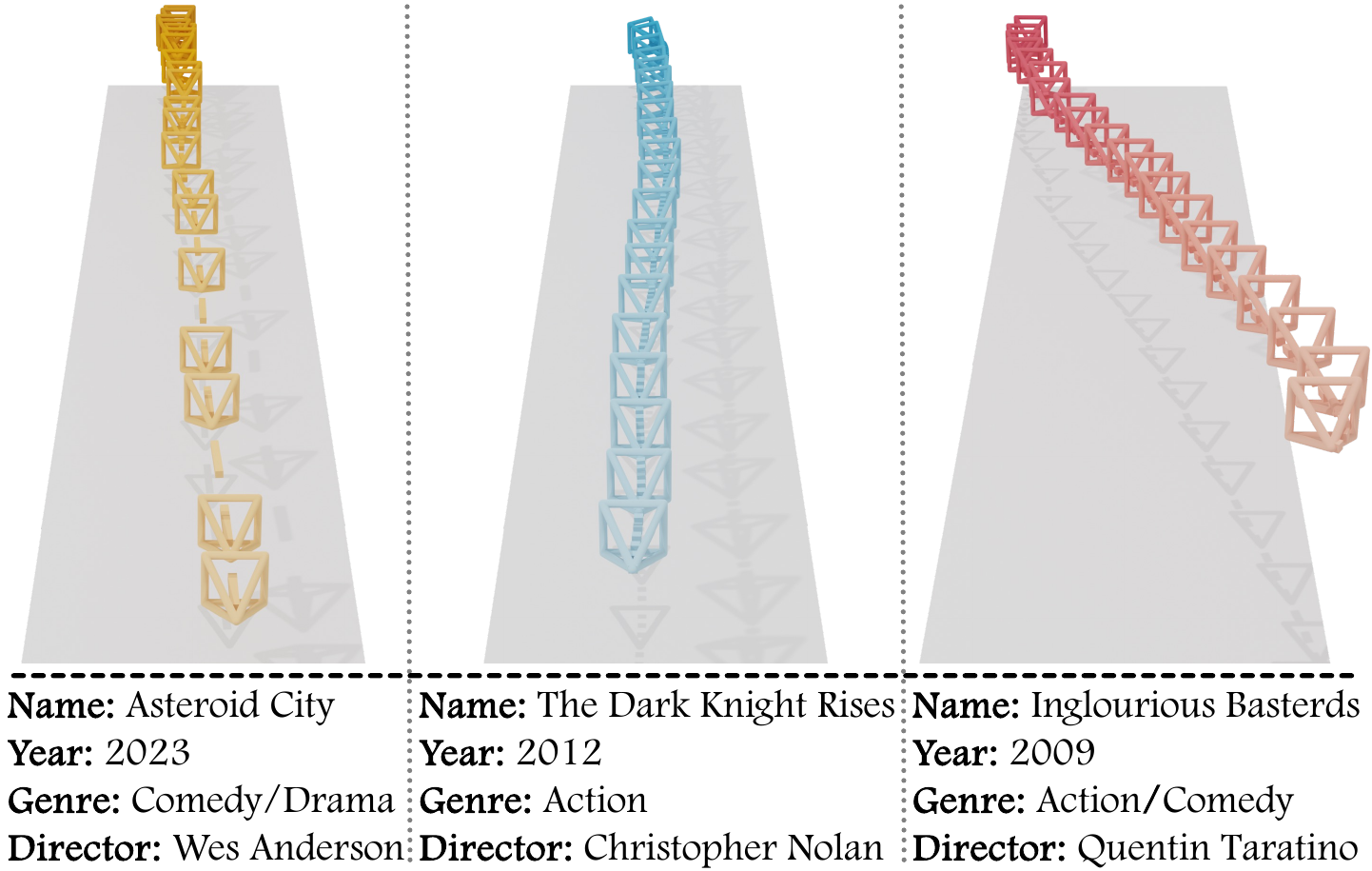}
    \caption{Dolly-in with different movie attributes}
    \label{fig:stat_traj_attr}
  \end{subfigure}
  \caption{Data statistics and examples in \dataset. 
  Left: distribution of extracted translation (yellow) and rotation  (green) primitives. 
  Right: examples of clips with a similar coarse motion pattern (dolly-in) but different movie attributes, 
  illustrating that camera style can vary substantially even under the same high-level motion. 
  }
  \label{fig:data_statistics}
\end{figure}

Figure~\ref{fig:data_statistics}(a) summarizes the distribution of the extracted motion primitives. While dominated by static and simple single-axis motions, which accurately reflects the natural imbalance of real cinematic camera behavior, the dataset still contains a substantial volume of multi-axis and composite motions to provide necessary diversity. Furthermore, Fig.~\ref{fig:data_statistics}(b) illustrates that clips sharing the same coarse motion pattern (e.g., a dolly-in) can exhibit markedly different trajectory styles depending on their movie attributes. This validates our motivation for incorporating metadata to explore cinematic structures beyond basic motion alignment. More detailed statistics are provided in Appendix~\ref{app:data_stats}.

\section{A Closer Look at Evaluation}
\label{sec:eval}

While decomposing camera motion into direction and speed is mathematically simple, this representational shift yields profound benefits for both trajectory alignment and generation. To accurately quantify these advantages, however, it is essential to first establish a reliable evaluation framework. Prior work commonly relies on CLaTr~\cite{courant2024et} for evaluation, which attempts to perform both text alignment and trajectory reconstruction simultaneously. This dual objective often obscures true alignment quality. To establish a clearer standard, we introduce a lightweight, purely contrastive evaluation protocol. Furthermore, leveraging \dataset, we complement standard text alignment with attribute-based probes to measure how well our representation captures higher-level cinematic information. The right panel of Fig.~\ref{fig:gen_eval_overview} summarizes this evaluation framework.

\subsection{Alignment Evaluation}
\label{sec:eval_alignment}

Unless otherwise specified, we evaluate alignment using motion captions, which provide the most direct text--trajectory correspondence. To project these modalities into a shared space, we employ a frozen CLIP ViT-L/14~\cite{radford2021clip} model as our text encoder $E_y$ and a lightweight Transformer as our trajectory encoder $E_x$. For a batch of $B$ matched pairs, let $\widehat{\mathbf{h}}^y_i = E_y(y_i)$ and $\widehat{\mathbf{h}}^x_i = E_x(x_i)$ denote the $L_2$-normalized embeddings for text and trajectory, respectively. Both CLaTr and our proposed protocol utilize a symmetric InfoNCE contrastive loss to align these representations:
$$ \mathcal{L}_{\mathrm{NCE}} = \frac{1}{2}(\mathcal{L}_{y\to x} + \mathcal{L}_{x\to y}), \quad \text{where} \quad \mathcal{L}_{y\to x} = -\frac{1}{B}\sum_{i=1}^{B} \log \frac{\exp(\langle \widehat{\mathbf{h}}^y_i,\widehat{\mathbf{h}}^x_i\rangle/\tau)}{\sum_{j\in\mathcal{N}(i)} \exp(\langle \widehat{\mathbf{h}}^y_i,\widehat{\mathbf{h}}^x_j\rangle/\tau)}. $$
Here, $\tau$ is a fixed temperature and $\mathcal{N}(i)$ contains the positive pair alongside non-duplicate negatives.

However, CLaTr~\cite{courant2024et} functions as a retrieval-VAE, coupling this contrastive loss with a heavy trajectory reconstruction objective. While this multi-objective design is useful for general representation learning, it compromises the precise discriminative margins required to evaluate strict instance-level alignment. To establish a more rigorous standard, we introduce a deterministic, encoder-only evaluation protocol. By stripping away the generative overhead and training solely with the contrastive objective ($\mathcal{L}_{\mathrm{NCE}}$), this lightweight setup strictly isolates pure alignment quality.

\begin{figure}[t]
  \centering
  \begin{subfigure}[t]{0.249\linewidth}
    \centering
    \includegraphics[width=\linewidth]{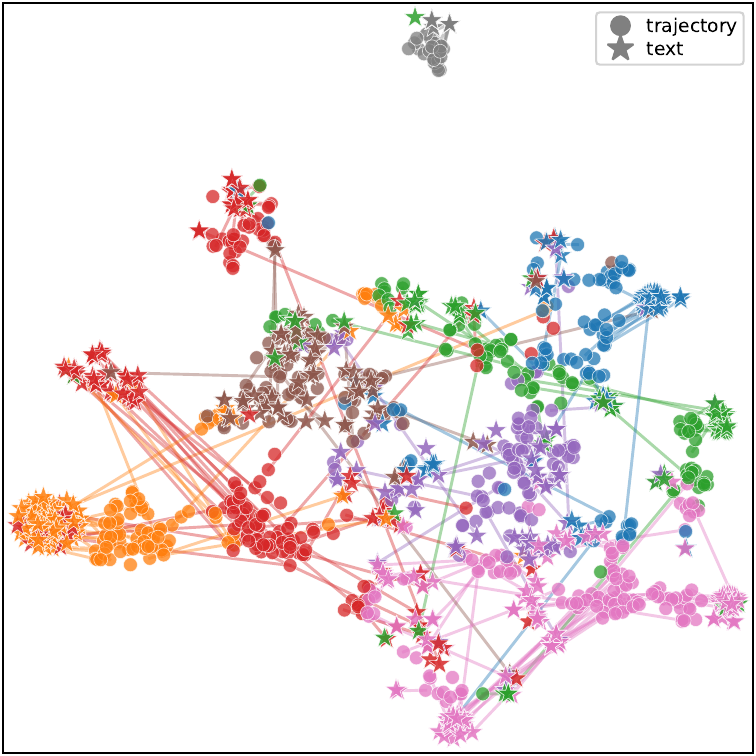}
    \caption{CLaTr-\dirspeed}
    \label{fig:joint_clatr_ds}
  \end{subfigure}\hfill
  \begin{subfigure}[t]{0.249\linewidth}
    \centering
    \includegraphics[width=\linewidth]{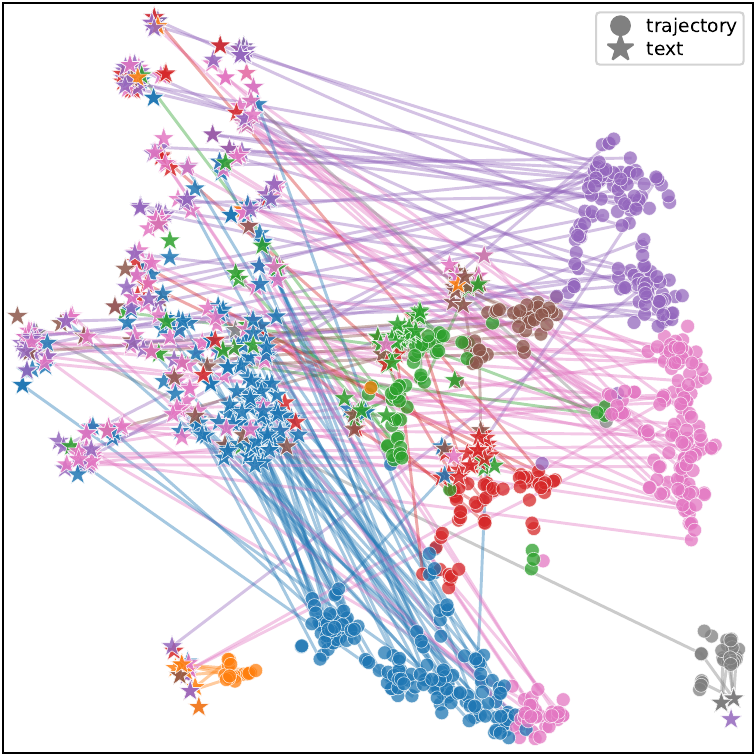}
    \caption{CLaTr-\pose}
    \label{fig:joint_clatr_pose}
  \end{subfigure}\hfill
  \begin{subfigure}[t]{0.249\linewidth}
    \centering
    \includegraphics[width=\linewidth]{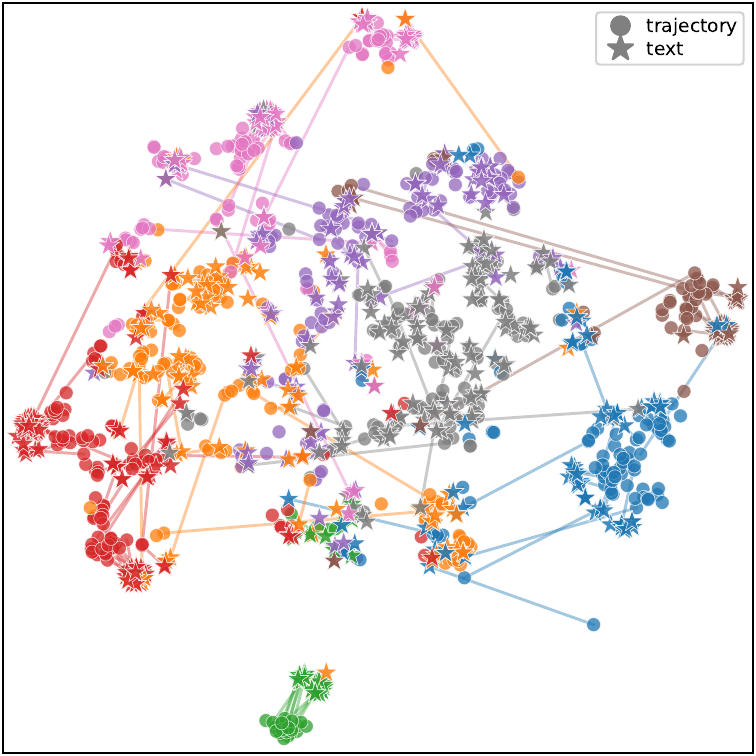}
    \caption{Ours-\dirspeed}
    \label{fig:joint_ours_ds}
  \end{subfigure}\hfill
  \begin{subfigure}[t]{0.249\linewidth}
    \centering
    \includegraphics[width=\linewidth]{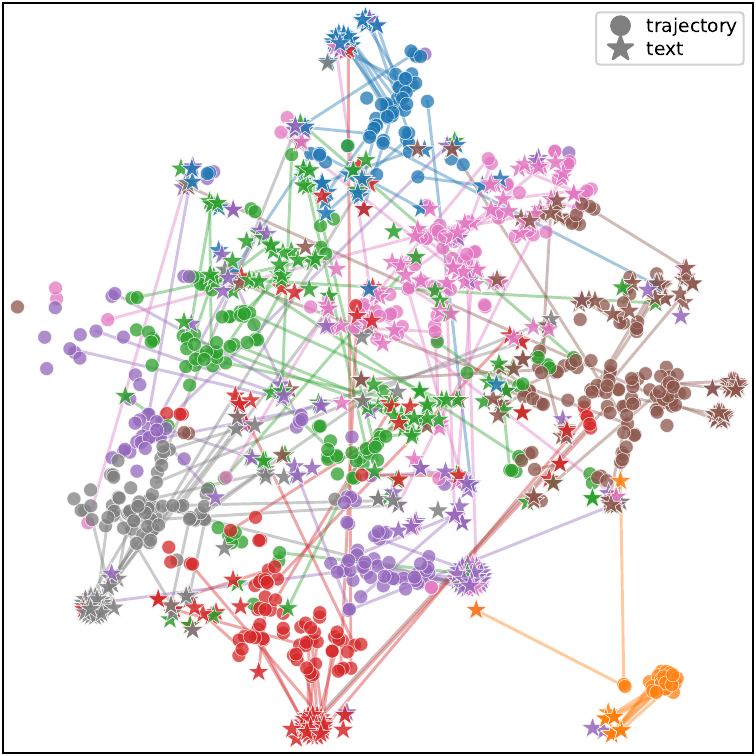}
    \caption{Ours-\pose}
    \label{fig:joint_ours_pose}
  \end{subfigure}
  \caption{Joint embedding spaces of the four alignment evaluators on the validation set, projected jointly with t-SNE. Trajectory embeddings ($\bullet$) and matched text embeddings ($\star$) are coloured by the same K-means cluster (computed on trajectory
   embeddings, $k=8$); thin lines connect $200$ random matched ($\bullet$, $\star$) pairs. 
  }
  \label{fig:joint_motion_4panel}
\end{figure}

Crucially, this clean evaluation space explicitly unveils the value of our motion-centric approach. As shown in Table~\ref{tab:align_eval}, our purely contrastive protocol significantly improves retrieval performance over CLaTr, with \dirspeed\ yielding a consistent and surprising boost regardless of the underlying evaluator (e.g., R@1 rising from $17.8$ to $25.2$). These gains are visually corroborated by the joint embedding spaces in Fig.~\ref{fig:joint_motion_4panel}. In the \dirspeed\ variants, trajectory points ($\bullet$) and matched text stars ($\star$) of the same cluster color are more integrated into cohesive cross-modal groups with minimal connecting lines. In contrast, the \pose\ variants exhibit disjoint color regions and numerous long lines, indicating that matched pairs remain far apart and poorly aligned. These results confirm that, despite its structural simplicity, mapping motion to direction and speed resolves geometric entanglement far more effectively than absolute poses.

\begin{table}[t]
  \centering\small
  \caption{Alignment evaluator validation on motion captions. We compare CLaTr and our purely contrastive protocol under both trajectory representations.}
  \label{tab:align_eval}
  \begin{tabular}{llcccccc}
    \toprule
    Model & Rep. & R@1\up & R@5\up & R@10\up & MedR\down & \alnscore\up & \#Params \\
    \midrule
    \multirow{2}{*}{Ours}
      & \dirspeed
        & \textbf{25.2} & \textbf{40.7} & \textbf{49.0} & \textbf{11} & \textbf{66.3} & 3.6\,M \\
      & \pose
        & 17.8 & 26.5 & 31.4 & 51 & 46.6 & 3.6\,M \\
    \midrule
    \multirow{2}{*}{CLaTr~\cite{courant2024et}}
      & \dirspeed
        & \textbf{19.7} & \textbf{30.5} & \textbf{39.2} & \textbf{23} & \textbf{70.6} & 30\,M \\
      & \pose
        & 6.9 & 12.0 & 14.7 & 361 & 37.4 & 30\,M \\
    \bottomrule
  \end{tabular}
\end{table}

\noindent\textbf{Metrics.}
To support both the immediate validation of our alignment space (Table~\ref{tab:align_eval}) and the subsequent evaluation of our generative model, we establish two sets of functional metrics. First, for \emph{text--trajectory alignment}, we compute the Alignment Score (\alnscore), defined concisely as the mean clipped cosine similarity $\frac{100}{N}\sum_{i} \max(0,\langle \widehat{\mathbf{h}}^y_i,\widehat{\mathbf{h}}^x_i\rangle)$, alongside retrieval recall (R@1, 5, 10) and Median Rank (MedR) in the trajectory-to-text direction. Second, for evaluating the physical realism and diversity of final generated outputs, we assess \emph{trajectory quality} using motion-primitive F1 (following the segmentation protocol of~\cite{courant2024et,zhang2025gendop}), Fr\'echet Camera Distance (\fcd) and Manifold Coverage~\cite{naeem2020prdc} computed in our new contrastive embedding space. Further implementation details are provided in Appendix~\ref{app:alignment_eval}.

\subsection{Attribute-based Evaluation}
\label{sec:eval_attribute}

Standard text alignment confirms adherence to local motion descriptions. However, as suggested by our earlier dataset statistics (Fig.~\ref{fig:data_statistics}(b)), the physical execution of a camera path encodes rich stylistic nuances that coarse motion captions simply cannot capture. To formally test whether a representation preserves these macro-level signatures, we train classifiers solely on real trajectories using the movie metadata in \dataset. 

Specifically, we construct three classification probes targeting Era, Genre, and Director. To mitigate the heavy long-tail distribution inherent in movie metadata, we group related genres into three broad buckets (Drama/Romance, Comedy, and Action/Thriller/Sci-Fi) and focus the directorial probe on three iconic stylists: Christopher Nolan, Wes Anderson, and Steven Spielberg. To ensure a rigorous evaluation, we restrict each task to these target categories, preventing classifiers from achieving artificially high performance by simply predicting a dominant background class. The Era probe follows a similar balanced design, formulated as a single-label task across three historical bins (Film, Early digital, and Digital mature). More details are previded in Appendix~\ref{app:attribute_eval}.

Table~\ref{tab:classifier_real} presents the performance of these attribute probes on real validation trajectories. While the absolute F1-scores suggest that these high-level attributes are not trivially mapped from motion alone, the results confirm that camera trajectories do carry a measurable stylistic signal. Directorial style, in particular, emerges as the most distinctive attribute among the three. Notably, \dirspeed\ consistently provides a more reliable signal than \pose\ across all tasks. This performance gap suggests that our decomposition helps expose subtle cinematic signatures that are otherwise difficult to capture through absolute coordinates. We leverage these real-trained probes to monitor attribute preservation in our generative experiments in Sec.~\ref{sec:experiments}.

\begin{table}[htbp]
  \centering\small
  \caption{Attribute prediction on real trajectories. All entries report macro-F1 ($\times 100$).}
  \label{tab:classifier_real}
  \begin{tabular}{lccccc}
    \toprule
    Attribute & Task & \#Class & \dirspeed & \pose & $\Delta$ \\
    \midrule
    Era      & single-label & 3 & \textbf{49.8} & 47.3 & $+2.6$ \\
    Genre    & multi-label  & 3 & \textbf{62.1} & 60.7 & $+1.4$ \\
    Director & single-label & 3 & \textbf{68.2} & 51.8 & $+16.4$ \\
    \bottomrule
  \end{tabular}
\end{table}

\section{Generation}
\label{sec:gen}

\begin{figure}[t]
  \centering
  \includegraphics[width=0.8\linewidth]{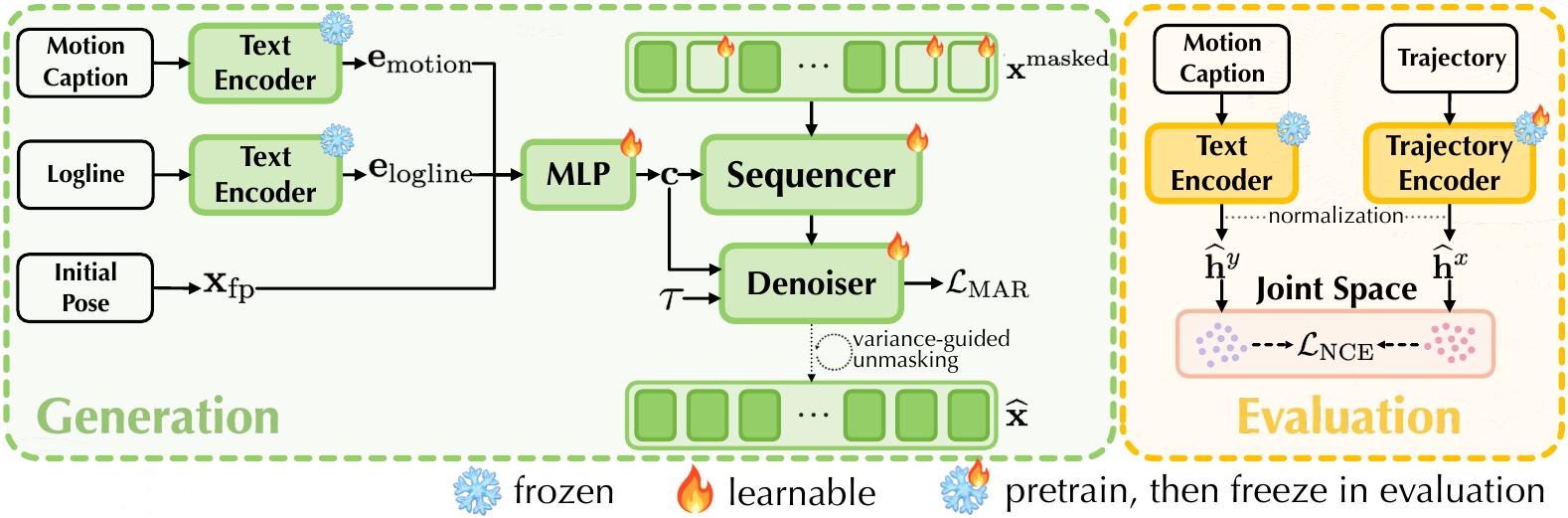}
  \caption{Overview of the proposed generation and evaluation framework. 
  Left: \ours\ integrates motion captions, loglines, and initial pose constraints to synthesize camera trajectories. The model progressively recovers masked trajectory tokens using a Transformer-based sequencer and a conditional denoiser through a variance-guided unmasking process. 
  Right: Our evaluation protocol employs a purely contrastive InfoNCE objective to align motion captions and trajectories within a normalized joint embedding space, providing a rigorous measure of cross-modal consistency.
  }
  \label{fig:gen_eval_overview}
\end{figure}

With the representation and evaluation protocol established, we now model the conditional distribution of continuous camera trajectories. Unlike image-generation tasks where Masked Auto-Regressive (MAR)~\cite{li2024mar} models typically operate on compressed latent tokens, our trajectory features are natively low-dimensional. We therefore design \ours\ to generate directly in the trajectory feature space, bypassing the need for a lossy autoencoding bottleneck. As illustrated in Fig.~\ref{fig:gen_eval_overview} (left), the model consists of two core components: a Transformer-based~\cite{vaswani2017attention} MAR sequencer that captures dependencies among partially visible trajectory tokens, and a conditional diffusion denoiser that samples the continuous values for missing positions.

\noindent\textbf{Masked sequence modeling.}
Let $\mathbf{x}=(\mathbf{x}_1,\ldots,\mathbf{x}_T)\in\mathbb{R}^{T\times D}$ represent a trajectory sequence. During training, we randomly sample a set of masked positions $\mathcal{M}\subseteq\{1,\ldots,T\}$ and replace them with a learned mask embedding $\mathbf{m}\in\mathbb{R}^{D}$:
$$ \mathbf{x}^{\mathrm{masked}}_i = \begin{cases} \mathbf{m}, & i\in\mathcal{M}, \\ \mathbf{x}_i, & i\notin\mathcal{M}. \end{cases} $$
The resulting sequence is processed by a Transformer-based sequencer $\mathrm{Seq}$, which employs self-attention to compute context vectors $\mathbf{H}^{\mathrm{seq}}$ from the visible tokens and conditioning signals:
$$ \mathbf{H}^{\mathrm{seq}} = \mathrm{Seq}(\mathbf{x}^{\mathrm{masked}},\mathbf{c}), \quad \mathbf{H}^{\mathrm{seq}}\in\mathbb{R}^{T\times h}. $$

\noindent\textbf{Text and first-pose conditioning.}
The conditioning vector $\mathbf{c}$ integrates three distinct signals: the motion caption, the logline, and the initial camera pose. We encode the two text components using a frozen CLIP ViT-B/32~\cite{radford2021clip} to obtain $\mathbf{e}_{\mathrm{motion}}$ and $\mathbf{e}_{\mathrm{logline}}$. 

The first-pose feature $\mathbf{x}_{\mathrm{fp}}$ ensures the generated motion remains geometrically grounded to the starting viewpoint. For the \pose\ representation, we use $\mathbf{x}^{\posemath}_1$ as defined in Sec.~\ref{sec:repr}. For \dirspeed, we construct an 8D first-pose feature by applying the same direction-speed decomposition to the initial camera state $(R_1, \mathbf{t}_1)$. These components are fused via a small MLP, $f_{\mathrm{cond}}$, to produce the final conditioning vector:
$ \mathbf{c} = f_{\mathrm{cond}}([\mathbf{e}_{\mathrm{motion}},\mathbf{e}_{\mathrm{logline}},\mathbf{x}_{\mathrm{fp}}])$.
This signal is injected into both the sequencer and the denoiser through adaptive layer normalization (AdaLN).

\noindent\textbf{Diffusion denoiser for continuous tokens.}
For each masked position $j\in\mathcal{M}$, we model the continuous value of $\mathbf{x}_j$ using a conditional diffusion process. Following the DDPM~\cite{ho2020denoising} forward process, we sample a diffusion step $\tau$ and Gaussian noise $\boldsymbol{\epsilon}$ to form a noised token $\mathbf{x}_{j,\tau}$. An MLP-based denoiser $\epsilon_\theta$ then predicts the noise conditioned on the sequencer's output:
$$ \widehat{\boldsymbol{\epsilon}} = \epsilon_\theta(\mathbf{x}_{j,\tau},\tau,\mathbf{H}^{\mathrm{seq}}_j). $$
The entire system is trained jointly to minimize the denoising error:
$$ \mathcal{L}_{\mathrm{MAR}} = \mathbb{E}_{j\in\mathcal{M},\,\tau,\,\boldsymbol{\epsilon}} \left[ \|\widehat{\boldsymbol{\epsilon}}-\boldsymbol{\epsilon}\|_2^2 \right]. $$

\noindent\textbf{Variance-guided unmasking order.}
At inference, \ours\ must determine the order in which masked positions are revealed. Rather than a fixed linear order, we employ a variance-guided policy. At each autoregressive step $s$, we compute the variance across the $h$ channels of the context vector $\mathbf{H}^{\mathrm{seq}}_j$ for each remaining masked position $j\in\mathcal{M}_s$. Let $H^{\mathrm{seq}}_{j,k}$ denote the $k$-th component of this vector; we score each position by:
$$ \rho_j = \frac{1}{h} \sum_{k=1}^{h} \left( H^{\mathrm{seq}}_{j,k} - \bar{H}^{\mathrm{seq}}_{j} \right)^2, \quad \text{where} \quad \bar{H}^{\mathrm{seq}}_{j} = \frac{1}{h}\sum_{k'=1}^{h} H^{\mathrm{seq}}_{j,k'}. $$
We then reveal the $n_s$ positions with the smallest variance, $\mathcal{P}_s = \operatorname*{arg\,min}_{\mathcal{P}\subseteq\mathcal{M}_s, |\mathcal{P}|=n_s} \sum_{j\in\mathcal{P}}\rho_j$, based on the intuition that lower variance indicates higher contextual confidence. This process repeats until the full trajectory is synthesized and converted back to per-frame poses.

\section{Experiments}
\label{sec:experiments}

We evaluate \ours\ against prior baselines on \dataset\ to quantify how the shift from traditional poses to our motion-centric (\dirspeed) representation improves generation quality. Unless otherwise stated, all metrics employ our contrastive evaluation protocol (Sec.~\ref{sec:eval}) using the \dirspeed\ encoder.

\paragraph{Setup and Baselines.}
We compare against three representative text-to-trajectory methods: \ccd~\cite{jiang2024ccd}, \director~\cite{courant2024et}, and \gendop~\cite{zhang2025gendop}. For each baseline, we evaluate both the original released pretrained checkpoints (denoted by $\dagger$) and versions retrained on \dataset\ to ensure a fair comparison. To avoid data leakage in our attribute-based evaluation, we train our movie-attribute classifiers on a separate movie-level split, ensuring that no clips from the same film are shared between classifier training and generation validation. All methods are evaluated using the metrics defined in Sec.~\ref{sec:eval}.

\begin{figure}[htbp]
  \centering
  \includegraphics[width=\linewidth]{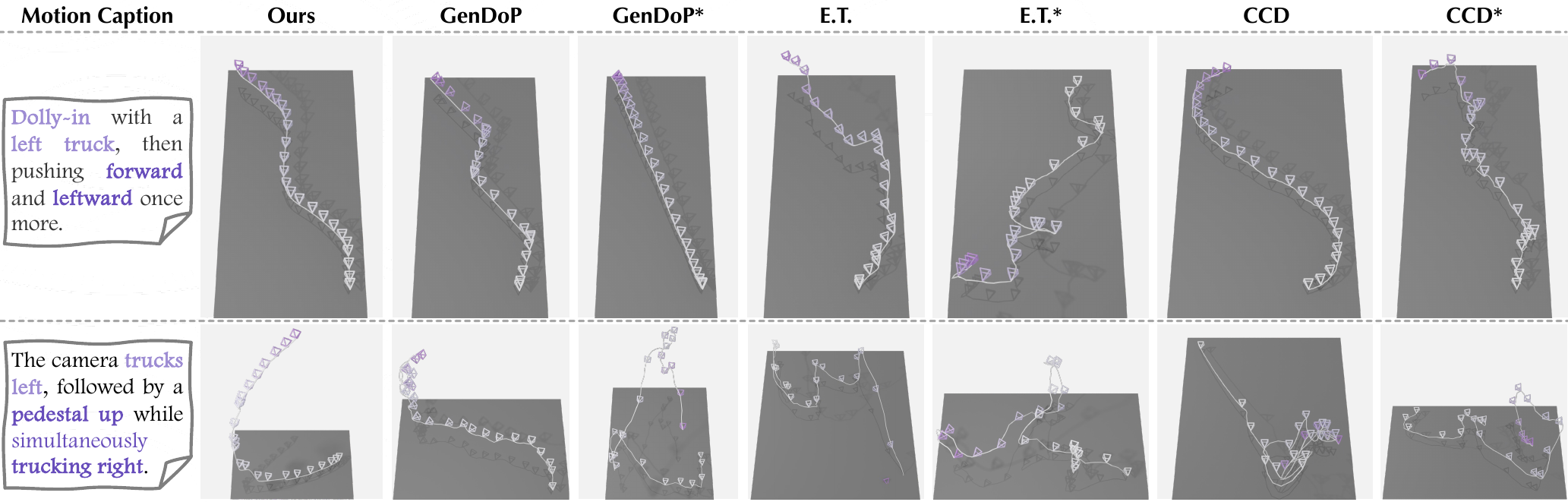}
  \caption{Qualitative comparison with prior generation methods. 
  Starred columns denote released pretrained checkpoints, corresponding to $*$ in Table~\ref{tab:gen_results}. 
  Compared with prior methods, \ours\ produces smoother and more coherent camera paths that better reflect compound motion instructions.}
  \label{fig:method_comparison}
\end{figure}

\paragraph{Comparison with Prior Methods.}
As Table~\ref{tab:gen_results} shows, \ours\ consistently outperforms all baselines across trajectory quality, text alignment, and cinematic attribute preservation. Compared to the strongest retrained baseline (\gendop), our model achieves a substantial leap in retrieval R@1 (from $0.93$ to $3.35$) while reducing \fcd\ by nearly $70\%$. Beyond local motion fidelity, \ours\ achieves the highest macro-F1 across all attribute probes, successfully capturing the macro-level cinematic structures identified in our dataset (Fig.~\ref{fig:data_statistics}). Supported by qualitative examples of smoother, more coherent compound paths (Fig.~\ref{fig:method_comparison}), these comprehensive gains demonstrate that generating directly in the \dirspeed\ space yields trajectories that are more realistic and precisely grounded.

Crucially, the advantages of our motion-centric approach extend beyond our specific architecture. As detailed in Appendix~\ref{app:extended_baselines}, retrofitting prior generative models with \dirspeed\ broadly improves their trajectory quality and alignment, demonstrating its utility as a strong inductive bias. However, representation alone is insufficient; state-of-the-art performance relies on the synergy between this parameterization and our \ours\ framework. Ablations (Appendix~\ref{app:ablations}) corroborate this: while reverting to absolute poses triggers the most severe degradation across all metrics, omitting the first-pose anchor or replacing variance-guided unmasking with a random schedule also leads to a substantial drop in overall trajectory quality. Furthermore, removing the auxiliary logline noticeably weakens both trajectory realism and movie-attribute preservation. While \dirspeed\ intrinsically ensures precise geometric alignment, Appendix~\ref{app:text_choice_alignment} qualitatively demonstrates that the logline provides the essential narrative anchor required for stylistically nuanced cinematic generation.

\begin{table}[t]
  \centering\small
  \caption{Comparison with prior text-to-trajectory methods on \dataset. $*$ denotes evaluation using released pretrained checkpoints, while other baseline rows are trained on \dataset. The light green row highlights our model. Bold indicates the best result.}
  \label{tab:gen_results}
  \resizebox{\linewidth}{!}{
  \begin{tabular}{lccccccccc}
    \toprule
    & \multicolumn{3}{c}{Trajectory Quality}
    & \multicolumn{3}{c}{Text--Trajectory Alignment}
    & \multicolumn{3}{c}{Movie Attributes} \\
    \cmidrule(lr){2-4}\cmidrule(lr){5-7}\cmidrule(lr){8-10}
    Method / Setting
    & F1\up & \fcd\down & Coverage\up
    & \alnscore\up & R@1\up & MedR\down
    & Era\up & Genre\up & Director\up \\
    \midrule
    \ccd$^*$~\cite{jiang2024ccd}
      & 0.117 & 52.45 & 0.298 & 4.90 & 0.08 & 1032.5 & 44.3 & 46.3 & 12.1 \\
    \ccd~\cite{jiang2024ccd}
      & 0.128 & 49.19 & 0.270 & 18.82 & 0.43 & 397.0 & 47.2 & 56.3 & 32.7 \\
    \director$^*$~\cite{courant2024et}
      & 0.015 & 135.74 & 0.034 & 0.00 & 0.08 & 960.0 & 40.7 & 56.7 & 25.0 \\
    \director~\cite{courant2024et}
      & 0.000 & 179.88 & 0.014 & 0.36 & 0.16 & 745.0 & 40.7 & 45.8 & 33.6 \\
    \gendop$^*$~\cite{zhang2025gendop}
      & 0.175 & 102.97 & 0.085 & 7.63 & 0.23 & 715.5 & 33.5 & 45.6 & 18.1 \\
    \gendop~\cite{zhang2025gendop}
      & 0.234 & 22.10 & 0.586 & 33.11 & 0.93 & 195.0 & 33.2 & 52.5 & 29.9 \\
    \rowcolor{oursrow}
    \textbf{\ours\ (ours)}
      & \textbf{0.437} & \textbf{6.77} & \textbf{0.783}
      & \textbf{57.79} & \textbf{3.35} & \textbf{52.0}
      & \textbf{47.8} & \textbf{60.7} & \textbf{48.1} \\
    \bottomrule
  \end{tabular}}
\end{table}

\section{Conclusion}

In this paper, we demonstrate that cinematic camera motion is fundamentally defined not just by \emph{where} the camera is positioned, but by \emph{how} it moves in terms of direction and speed. By decomposing the camera trajectory into these components (\dirspeed), we unlock surprising benefits for both trajectory alignment and generation. Beyond its numerical advantages, this motion-centric approach inherently allows models to capture the higher-level cinematographic information that traditional absolute poses obscure. To support and rigorously evaluate and validate this insight, we introduced \dataset\ for scene-level and attribute-based analysis, a purely contrastive alignment protocol, and \ours, a tailored masked autoregressive generator. Together, these contributions overcome the limitations of prior pose-centric baselines, establishing a new standard for synthesizing realistic and precisely text-aligned camera paths. Future work will leverage this motion-centric foundation to explore richer multimodal conditioning and dedicated architectures for controllable cinematic styling.

\newpage


\bibliographystyle{unsrtnat}
\bibliography{refs}

\newpage

\appendix
\section*{Appendix}

This appendix provides supplementary details, extended experimental results, and ablation studies to support the core findings in the main text. We begin by detailing the logline generation pipeline and additional dataset statistics. We then formalize the evaluation metrics and the attribute-based evaluation protocol. Finally, we provide an extended experiments and an empirical analysis of text choice for alignment evaluation.

\section{Logline Generation Details}
\label{app:logline_generation}

\noindent\textbf{Motivation.}
Motion captions describe local camera behavior, such as panning, dollying, or tracking. While they provide direct supervision for motion-level controllability, they ignore the visual and narrative context in which the motion occurs. To address this, we associate each clip with a screenplay-style logline that summarizes the scene content. The goal of the logline is not to prescribe a strict, unique camera trajectory, but rather to provide scene-level context for studying text-to-trajectory generation.

\noindent\textbf{Generation procedure.}
We generate loglines using Qwen3-VL-32B-Instruct~\cite{qwen3vl}. The vision-language model takes the video clip as input and is prompted to summarize the visible scene in a concise, screenplay-style format. Crucially, the prompt instructs the model to focus on the setting, time of day, main subject, and visible action, while strictly avoiding any camera-motion descriptions. 

\noindent\textbf{Prompt used for annotation.}
The following prompt illustrates the exact instruction used in our pipeline:
\begin{quote}
\small
\textbf{Role:} You are a professional Assistant Director and Script Supervisor.

\textbf{Task:} Analyze the provided video clip and write a concise screenplay-style logline describing the scene.

\textbf{Goal:} Produce a short scene description that captures the setting, time of day, main subject, and visible action. The description should summarize what is happening in the video, not how the camera moves.

\textbf{Constraints:}
\begin{enumerate}
    \item Do not mention camera motion, camera direction, zoom, pan, tilt, dolly, or tracking.
    \item Use present tense.
    \item Keep the description concise and visually grounded.
    \item Avoid hallucinating details that are not visible or strongly implied by the video.
    \item Use standard screenplay-style format.
\end{enumerate}

\textbf{Output format:}

\texttt{[INT./EXT.] [Specific Location] - [Time of Day] - [Brief Action/Subject Description]}

\textbf{Examples:}

\texttt{INT. HOSPITAL CORRIDOR - NIGHT - A nurse runs toward the emergency room.}

\texttt{EXT. MOUNTAIN RIDGE - DAY - Clouds roll over the jagged peaks.}

\textbf{Output only the logline.}
\end{quote}

\noindent\textbf{Output format.}
The final logline follows a standard screenplay-inspired format, e.g.:
\begin{quote}
\emph{EXT. STONE BRIDGE OVER RIVER -- DAY -- A lone hiker in red crosses a stone bridge under a clear sky.}
\end{quote}
The descriptions are consistently kept in the present tense. 

\noindent\textbf{Quality control.}
We manually review and correct generated loglines to ensure consistency and quality. Specifically, we fix malformed screenplay formats, remove hallucinated details that conflict with the visual evidence, and condense overly verbose descriptions. This manual review ensures that the loglines reliably serve as conditioning inputs for studying the relationship between scene context and camera motion.

\noindent\textbf{Relation to motion captions.}
Motion captions and loglines provide complementary supervision. A motion caption dictates the camera action directly (e.g., \emph{``the camera slowly dollies in''}). A logline instead establishes the narrative context (e.g., \emph{``INT. DIMLY LIT ROOM -- NIGHT -- A detective studies a wall of photographs''}). Consequently, motion captions provide a relatively strict text--trajectory correspondence, whereas loglines are inherently underdetermined—many plausible camera trajectories might suit the same scene. For this reason, we rely on motion captions as the primary text for alignment evaluation, while utilizing loglines as an auxiliary conditioning signal for generation.

\section{Additional Dataset Statistics}
\label{app:data_stats}

\begin{figure*}[htbp]
  \centering
  \includegraphics[width=\linewidth]{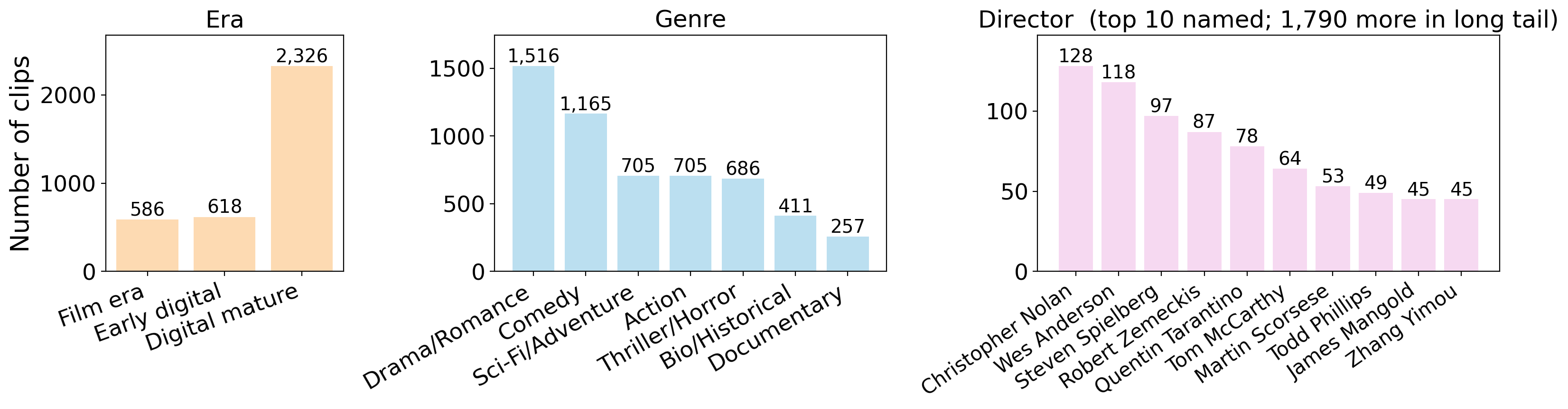}
  \caption{Distribution of linked movie attributes in \dataset. 
  Left: era distribution. 
  Middle: genre distribution. 
  Right: director distribution, highlighting only the top 10 named directors alongside the grouped long tail. 
  This inherent imbalance motivates our use of targeted, frequent-class subsets for rigorous attribute evaluation.}
  \label{fig:attr_dist}
\end{figure*}

\noindent\textbf{Movie attribute distribution.}
A subset of \dataset\ can be explicitly linked to real movie metadata, enabling the attribute-based analyses presented in the main paper. Figure~\ref{fig:attr_dist} visualizes the distribution of this linked subset across era, genre, and director labels. The era labels are historically imbalanced, with a heavy concentration in the ``Digital mature'' period. Genre labels exhibit more diversity but remain skewed toward several dominant categories. Director labels show a severe long-tail distribution: even the most frequently credited directors account for only a small fraction of the total dataset, with thousands of others appearing only rarely. To mitigate this heavy long-tail effect during evaluation, we restrict our attribute-based probes to frequent-class subsets, as detailed in Sec.~\ref{sec:eval_attribute}.

\noindent\textbf{Comparison with prior datasets.}
Table~\ref{tab:dataset_compare} compares \dataset\ against representative datasets for text-to-camera-trajectory generation. While comparable in raw scale to recent movie-based datasets, \dataset\ is uniquely distinguished by its annotation richness. By jointly providing motion captions, loglines, and a metadata-linked subset with real movie attributes, it uniquely supports both motion- and scene-conditioned generation, alongside high-level cinematographic evaluation.

\begin{table}[htbp]
  \centering\small
  \caption{Comparison of datasets for text-to-camera-trajectory generation. 
\dataset\ is distinguished by the joint availability of motion captions, loglines, and linked real movie attributes.}
  \label{tab:dataset_compare}
  \resizebox{\linewidth}{!}{
  \begin{tabular}{l r r r ccc l}
    \toprule
    & \multicolumn{3}{c}{Scale} & \multicolumn{3}{c}{Annotations} & \\
    \cmidrule(lr){2-4}\cmidrule(lr){5-7}
    Dataset & \#Samples & \#Frames & Avg(s) & motion & logline & movie attributes & Source \\
    \midrule
    CCD~\cite{jiang2024ccd}
      & 25\,K & 4.5\,M & 7.2
      & \cmark & \xmark & \xmark & Synthetic \\
    E.T.~\cite{courant2024et}
      & 115\,K & 11\,M & 3.8
      & \cmark & \xmark & \xmark & Movie \\
    DataDoP~\cite{zhang2025gendop}
      & 29\,K & 11\,M & 14.4
      & \cmark & \xmark & \xmark & Movie \\
    \midrule
    \textbf{\dataset\ (ours)}
      & 28\,K & 10\,M & 12.0
      & \cmark & \cmark & \cmark\,($3.2$K) & Movie \\
    \bottomrule
  \end{tabular}}
\end{table}

\section{Alignment Evaluation Metrics}
\label{app:alignment_eval}

This section formalizes the text--trajectory alignment and trajectory-quality metrics utilized in our evaluation framework.

\subsection{Text--Trajectory Alignment Metrics}

Unless otherwise specified, alignment is evaluated using the motion caption paired with each trajectory. For a validation set of $N$ matched pairs $\{(\mathbf{x}_i,\mathbf{y}_i)\}_{i=1}^{N}$, we extract normalized trajectory and text embeddings:
$$ \widehat{\mathbf{h}}^x_i = \frac{E_x(\mathbf{x}_i)}{\|E_x(\mathbf{x}_i)\|_2}, \qquad \widehat{\mathbf{h}}^y_i = \frac{E_y(\mathbf{y}_i)}{\|E_y(\mathbf{y}_i)\|_2}, $$
where $E_x$ and $E_y$ are the trajectory and text encoders, respectively. The resulting cosine-similarity matrix is:
$$ S_{ij} = \left\langle \widehat{\mathbf{h}}^x_i, \widehat{\mathbf{h}}^y_j \right\rangle, \qquad S\in\mathbb{R}^{N\times N}, $$
where the diagonal entry $S_{ii}$ represents the similarity of the matched pair.

\noindent\textbf{Alignment score.}
The alignment score (\alnscore) computes the average clipped cosine similarity of all matched pairs:
$$ \mathrm{\alnscore} = \frac{100}{N} \sum_{i=1}^{N} \max\left(0,S_{ii}\right). $$
While this captures absolute matched-pair similarity, it does not evaluate discrimination against negative distractors. Because separately trained evaluators define different embedding spaces, their absolute \alnscore\ values are not directly comparable; we use retrieval metrics for cross-evaluator comparisons.

\noindent\textbf{Retrieval recall.}
For trajectory-to-text retrieval, each trajectory embedding $\widehat{\mathbf{h}}^x_i$ acts as a query, and all text embeddings $\{\widehat{\mathbf{h}}^y_j\}_{j=1}^{N}$ are ranked by descending similarity $S_{ij}$. Let $\mathrm{rank}_i$ be the $1$-indexed rank of the correct text $\mathbf{y}_i$. Ignoring ties, this is defined as:
$$ \mathrm{rank}_i = 1+ \sum_{j\neq i} \mathbf{1}\left[S_{ij}>S_{ii}\right], $$
where $\mathbf{1}[\cdot]$ is the indicator function. The Recall@$K$ metric is then:
$$ \mathrm{R@}K = \frac{100}{N} \sum_{i=1}^{N} \mathbf{1}\left[\mathrm{rank}_i\leq K\right], \qquad K\in\{1,5,10\}. $$
In practice, ties are resolved by assigning the average rank among tied candidates.

\noindent\textbf{Median rank.}
The Median Rank (MedR) summarizes the overall retrieval distribution:
$$ \mathrm{MedR} = \mathrm{median}\left(\mathrm{rank}_1,\ldots,\mathrm{rank}_N\right). $$
A lower MedR indicates that, on average, the correct match is ranked closer to the top. The text-to-trajectory direction is computed symmetrically using $S^\top$. As noted in the main text, trajectory-to-text retrieval serves as our primary metric, since it directly answers whether the text conditioning is recoverable from the generated physical motion.

\subsection{Trajectory-Quality Metrics}

For physical quality evaluation, each camera trajectory is treated as a sequence of camera-to-world matrices $P_{1:T}=(P_1,\ldots,P_T)$, where $P_t\in SE(3)$.

\noindent\textbf{Motion-primitive F1.}
This metric evaluates instance-level structural fidelity. For each consecutive frame pair, we extract the relative transformation:
$$ \Delta_t = P_t^{-1}P_{t+1}, \qquad t=1,\ldots,T-1. $$
From $\Delta_t$, we isolate the translational velocity $\mathbf{v}_t\in\mathbb{R}^3$ and angular velocity $\boldsymbol{\omega}_t\in\mathbb{R}^3$. We quantize the translation into 27 sign-based bins ($\{-1,0,+1\}^3$) and the rotation into 7 bins (stationary plus positive/negative rotation around each dominant axis). This yields a 189-class label $\ell_t\in\{0,\ldots,188\}$ for each frame transition. Following prior protocols, the label sequence is smoothed via mode filtering. 

Let $\mathrm{Prec}_c$ and $\mathrm{Rec}_c$ represent the precision and recall for class $c$, and $n_c$ the number of reference frames belonging to $c$. The weighted multi-class F1 is:
$$ \mathrm{F1}_c = \frac{2\,\mathrm{Prec}_c\,\mathrm{Rec}_c}{\mathrm{Prec}_c+\mathrm{Rec}_c}, \qquad \mathrm{F1} = \frac{\sum_c n_c\,\mathrm{F1}_c}{\sum_c n_c}. $$

\noindent\textbf{Fr\'echet Camera Distance.}
\fcd\ measures the distributional gap between real and generated trajectories within the learned feature space of our contrastive trajectory encoder. Let $\phi(\cdot)$ denote the unnormalized embedding. For sets of real ($X_r$) and generated ($X_g$) trajectories, we compute embeddings $\mathbf{u}^{r}_i=\phi(\mathbf{x}^{r}_i)$ and $\mathbf{u}^{g}_j=\phi(\mathbf{x}^{g}_j)$. Fitting Gaussian statistics to these sets yields moments $(\boldsymbol{\mu}_r,\boldsymbol{\Sigma}_r)$ and $(\boldsymbol{\mu}_g,\boldsymbol{\Sigma}_g)$:
$$ \boldsymbol{\mu} = \frac{1}{N}\sum_i \mathbf{u}_i, \qquad \boldsymbol{\Sigma} = \frac{1}{N-1}\sum_i (\mathbf{u}_i-\boldsymbol{\mu})(\mathbf{u}_i-\boldsymbol{\mu})^\top. $$
The distance is then defined as:
$$ \mathrm{\fcd} = \|\boldsymbol{\mu}_r-\boldsymbol{\mu}_g\|_2^2 + \mathrm{Tr}(\boldsymbol{\Sigma}_r) + \mathrm{Tr}(\boldsymbol{\Sigma}_g) - 2\,\mathrm{Tr}\left((\boldsymbol{\Sigma}_r\boldsymbol{\Sigma}_g)^{1/2}\right). $$

\noindent\textbf{Manifold Coverage.}
Coverage assesses the proportion of the real trajectory manifold successfully reached by generated samples, again utilizing the contrastive trajectory encoder. We compute $L_2$-normalized embeddings:
$$ \bar{\mathbf{u}}^{r}_i = \frac{\phi(\mathbf{x}^{r}_i)}{\|\phi(\mathbf{x}^{r}_i)\|_2}, \qquad \bar{\mathbf{u}}^{g}_j = \frac{\phi(\mathbf{x}^{g}_j)}{\|\phi(\mathbf{x}^{g}_j)\|_2}. $$
For each real embedding $\bar{\mathbf{u}}^{r}_i$, we calculate the distance to its $k$-th nearest real neighbor ($k=3$):
$$ \rho_i = \left\| \bar{\mathbf{u}}^{r}_i - \bar{\mathbf{u}}^{r}_{i,(k)} \right\|_2. $$
A real sample is considered ``covered'' if at least one generated sample falls within this local radius:
$$ \mathrm{cov}_i = \mathbf{1}\left[ \min_{j} \left\| \bar{\mathbf{u}}^{r}_i - \bar{\mathbf{u}}^{g}_j \right\|_2 < \rho_i \right]. $$
The final Coverage score averages this indicator across the real set:
$$ \mathrm{Cov} = \frac{1}{N_r} \sum_{i=1}^{N_r} \mathrm{cov}_i. $$

\section{Details of Attribute-based Evaluation}
\label{app:attribute_eval}

\noindent\textbf{Task formulation.}
We utilize movie metadata linked from public knowledge bases to construct the three attribute probes (Era, Genre, and Director) summarized in Table~\ref{tab:attribute_task_defs}. Era and Director are formulated as single-label classification tasks. Given that modern films frequently span multiple stylistic categories, Genre is treated as a multi-label classification task.

\begin{table}[htbp]
  \centering\small
  \caption{Attribute task definitions. Clips outside the target classes are completely dropped rather than assigned to a generic ``Other'' class. This strictly prevents the classifiers from cheating the metric by defaulting to a trivial majority-class background prediction.}
  \label{tab:attribute_task_defs}
  \setlength{\tabcolsep}{4pt}
  \begin{tabular}{lll}
    \toprule
    Attribute & Classes & Type \\
    \midrule
    Era
      & Film era ($<2005$); Early digital ($2005$--$2012$); Digital mature ($\geq 2013$)
      & single-label \\
    Genre
      & Drama/Romance; Comedy; Action/Thriller/Sci-Fi
      & multi-label \\
    Director
      & Christopher Nolan; Wes Anderson; Steven Spielberg
      & single-label \\
    \bottomrule
  \end{tabular}
\end{table}

\noindent\textbf{Genre coarsening.}
To handle the extreme granularity and noise in Wikidata genre labels, we map them into coarse groups and merge related themes into three robust buckets. Drama and Romance are merged into one bucket, while action-oriented genres (Action, Thriller/Horror, Sci-Fi/Fantasy, Adventure) are collapsed into another. This prevents arbitrary decision boundaries among overlapping themes. Clips containing exclusively documentary or biographical labels are discarded for this specific probe.

\noindent\textbf{Classifier architecture.}
The attribute classifier consists of a small Transformer encoder processing the trajectory features, fused with a trajectory-statistics vector, a first-pose summary, and a frozen CLIP feature extracted from the first frame's depth map. Separate probes are trained from scratch for \dirspeed\ and \pose\ to ensure fair comparison. 

\noindent\textbf{Evaluation on generated trajectories.}
These probes measure preservation of latent attribute correlations rather than explicit controllable style, since the generators are not conditioned on era, genre, or director labels. The classifiers are trained on a separate movie-level split, so clips from the same film do not appear in both classifier training and validation. For generated trajectories, every compared method is evaluated with the same frozen classifier and identical first-pose/depth auxiliary inputs; only the generated trajectory differs across methods. We evaluate all generated validation clips whose corresponding source clips possess defined labels for a given task. Generated trajectories are never used to train the classifiers.

\section{Extended Baseline Comparison}
\label{app:extended_baselines}

Table~\ref{tab:extended_baselines} expands the main comparison with diagnostic variants, denoted by \dirspeedmark, that replace each baseline's native trajectory output with continuous \dirspeed\ prediction while retaining the rest of its generative framework. These variants isolate how well the representation transfers across architectures. As in the main paper, \pretrainedmark\ denotes released pretrained checkpoints, while unmarked rows denote versions retrained on \dataset\ using each method's native formulation. The retrained \director\ row uses its complete released training and sampling pipeline, including EDM preconditioning, EMA, and valid-length masking.

The diagnostic variants show that \dirspeed\ generally benefits motion F1 and language alignment, but the gains are not uniform across all distributional metrics. In particular, \gendop\ was originally designed to predict categorical distributions over discrete pose tokens; replacing this output space with continuous motion changes the demands placed on the same causal autoregressive framework and can affect \fcd\ and Coverage. This motivates pairing the continuous representation with a generator designed for it, rather than interpreting representation changes independently of architecture.

\begin{table}[htbp]
  \centering\small
  \caption{Extended comparison against prior generation methods on \dataset. \pretrainedmark\ denotes released pretrained checkpoints; unmarked rows denote baselines retrained on \dataset\ using their native formulations; and \dirspeedmark\ denotes diagnostic variants retrained with continuous \dirspeed\ prediction and motion-caption conditioning. Movie-attribute metrics report macro-F1 ($\times 100$).}
  \label{tab:extended_baselines}
  \resizebox{\linewidth}{!}{
  \begin{tabular}{lccccccccc}
    \toprule
    & \multicolumn{3}{c}{Trajectory Quality}
    & \multicolumn{3}{c}{Text--Trajectory Alignment}
    & \multicolumn{3}{c}{Movie Attributes} \\
    \cmidrule(lr){2-4}\cmidrule(lr){5-7}\cmidrule(lr){8-10}
    Method
    & F1\up & \fcd\down & Coverage\up
    & \alnscore\up & R@1\up & MedR\down
    & Era\up & Genre\up & Director\up \\
    \midrule
    \ccd\pretrainedmark~\cite{jiang2024ccd}
      & 0.117 & 52.45 & 0.298 & 4.90 & 0.08 & 1032.5 & 44.3 & 46.3 & 12.1 \\
    \ccd~\cite{jiang2024ccd}
      & 0.128 & 49.19 & 0.270 & 18.82 & 0.43 & 397.0 & 47.2 & 56.3 & 32.7 \\
    \rowcolor{diagrow}
    \ccd\dirspeedmark~\cite{jiang2024ccd}
      & 0.168 & 25.96 & 0.576 & 11.71 & 0.35 & 690.0 & 42.5 & 48.5 & 39.8 \\
    \midrule
    \director\pretrainedmark~\cite{courant2024et}
      & 0.015 & 135.74 & 0.034 & 0.00 & 0.08 & 960.0 & 40.7 & 56.7 & 25.0 \\
    \director~\cite{courant2024et}
      & 0.194 & 16.39 & 0.596 & 29.78 & 0.81 & 251.0 & 42.0 & 57.5 & 28.1 \\
    \rowcolor{diagrow}
    \director\dirspeedmark~\cite{courant2024et}
      & 0.338 & 48.32 & 0.277 & 25.22 & 0.19 & 389.5 & 40.0 & 50.2 & 37.2 \\
    \midrule
    \gendop\pretrainedmark~\cite{zhang2025gendop}
      & 0.175 & 102.97 & 0.085 & 7.63 & 0.23 & 715.5 & 33.5 & 45.6 & 18.1 \\
    \gendop~\cite{zhang2025gendop}
      & 0.234 & 22.10 & 0.586 & 33.11 & 0.93 & 195.0 & 33.2 & 52.5 & 29.9 \\
    \rowcolor{diagrow}
    \gendop\dirspeedmark~\cite{zhang2025gendop}
      & 0.293 & 55.30 & 0.260 & 40.62 & 0.93 & 124.0 & 40.5 & 48.4 & 30.6 \\
    \midrule
    \rowcolor{oursrow}
    \textbf{\ours\ (Ours)}
      & \textbf{0.437} & \textbf{6.77} & \textbf{0.783}
      & \textbf{57.79} & \textbf{3.35} & \textbf{52.0}
      & \textbf{47.8} & \textbf{60.7} & \textbf{48.1} \\
    \bottomrule
  \end{tabular}}
\end{table}

\paragraph{Conditioning-matched comparison}
\label{app:baseline_fairness}
The native baselines use motion-caption conditioning, whereas the full \ours\ model additionally uses the scene logline and first-pose anchor. Adding these modules to the baselines would substantially alter their original architectures, so we retain their released conditioning interfaces and instead remove both auxiliary conditions from \ours\ for a fully matched caption-only comparison.

\begin{table}[htbp]
\centering\small
\caption{Conditioning-matched comparison. The caption-only \ours\ variant removes both logline and first-pose conditioning.}
\label{tab:caption_only}
\resizebox{\linewidth}{!}{
\begin{tabular}{lccccccccc}
\toprule
Method & F1\up & \fcd\down & Coverage\up & \alnscore\up & R@1\up & MedR\down & Era\up & Genre\up & Director\up \\
\midrule
\ours & 0.437 & 6.77 & 0.783 & 57.79 & 3.35 & 52 & 47.8 & 60.7 & 48.1 \\
\ours\ \emph{caption-only} & 0.419 & 6.76 & 0.749 & 65.83 & 5.82 & 42 & 45.6 & 60.6 & 41.6 \\
\gendop~\cite{zhang2025gendop} & 0.234 & 22.10 & 0.586 & 33.11 & 0.93 & 195 & 33.2 & 52.5 & 29.9 \\
\bottomrule
\end{tabular}}
\end{table}

The full and caption-only \ours\ variants achieve comparable trajectory quality, while the caption-only model scores higher on caption-based alignment because it focuses exclusively on the signal used by those metrics. Under the matched caption-only setting, \ours\ still outperforms \gendop\ across all reported metrics, showing that richer conditioning does not explain the main performance gap.

\section{Human Evaluation}
\label{app:human_eval}

We conduct a blinded multi-selection study on \NumHumanShots{} randomly sampled
CameraBench test clips. For each clip, the source scene and the camera-conditioned
renderer are fixed; only the generated trajectory changes. We generate one trajectory
from each of \NumHumanMethods{} methods, re-anchor it to the source pose, and rescale
its translation extent to match the reference. Participants view the reference and
seven anonymized re-renderings and select every version whose camera motion is smooth,
natural, and follows the reference direction. Multiple selections are allowed, and
the same rendering settings and random seed are used for all methods.

\NumHumanParticipants{} participants completed all clips, yielding
\NumHumanJudgements{} method-level judgements. One participant was an author; removing
that response changes the selection rate by only $0.2$ percentage points and leaves
the conclusions unchanged. We report the selection rate (the fraction of judgements
in which a method is selected) and the stricter sole-selection rate (the fraction in
which it is the only selected method). Since several methods can be selected for one
clip, selection rates do not sum to $100\%$.

\begin{figure*}[t]
  \centering
  \includegraphics[width=\linewidth]{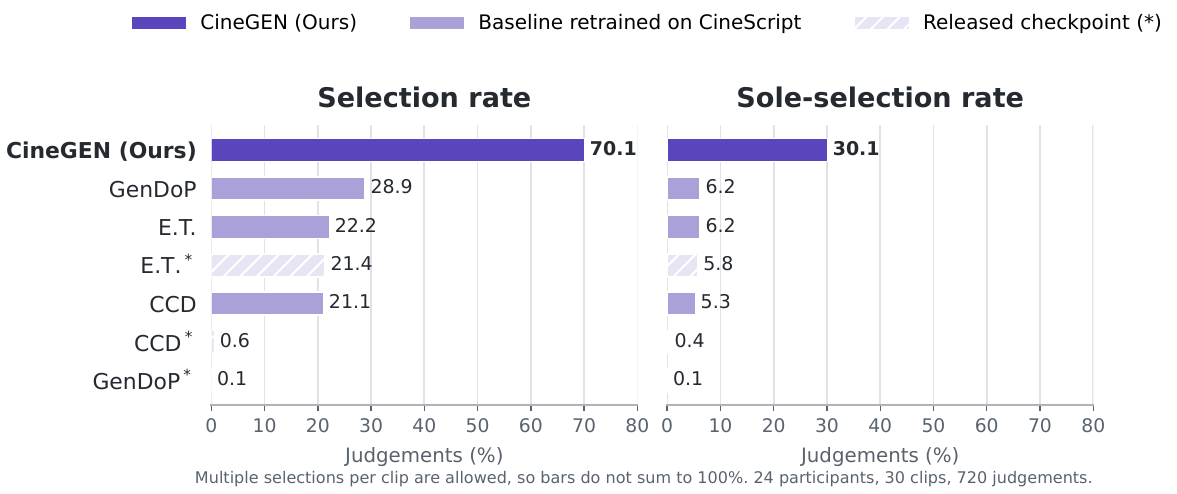}
  \caption{Blinded human evaluation across \NumHumanShots{} clips and
  \NumHumanParticipants{} participants. Multiple selections are allowed, so the
  selection-rate bars do not sum to $100\%$.}
  \label{fig:human_eval_rates}
\end{figure*}

Figure~\ref{fig:human_eval_rates} summarizes the
results. \ours\ is selected in $70.1\%$ of judgements and is the sole selection in
$30.1\%$, compared with $28.9\%$ and $6.2\%$ for the strongest baseline,
the retrained \gendop. Every participant selected \ours\ more often than that
baseline individually (two-sided sign test, $p<10^{-6}$). The released GenDoP and CCD
models are selected in less than $1\%$ of judgements, while their re-implemented
counterparts recover part of the gap, supporting the importance of the trajectory
representation in this comparison.

\section{Evaluator Robustness and Retrieval Context}
\label{app:evaluator_robustness}

\paragraph{Independent frozen-CLaTr evaluation.}
To ensure that the main generation results do not depend on our \dirspeed-trained evaluation space, we re-evaluate all methods using the frozen official CLaTr encoder released with E.T.. Table~\ref{tab:independent_clatr} shows that \ours\ remains best on all metrics under this independent embedding space.

\begin{table}[htbp]
\centering\small
\caption{Independent evaluation using the frozen official CLaTr encoder. \pretrainedmark\ denotes released pretrained checkpoints, while unmarked baselines are retrained on \dataset\ using their native formulations.}
\label{tab:independent_clatr}
\begin{tabular}{lcccc}
\toprule
Method & \fcd\down & Coverage\up & R@1\up & MedR\down \\
\midrule
\ccd\pretrainedmark~\cite{jiang2024ccd} & 195.23 & 0.293 & 0.39 & 1158.5 \\
\ccd~\cite{jiang2024ccd} & 567.67 & 0.253 & 0.43 & 965.5 \\
\director\pretrainedmark~\cite{courant2024et} & 1311.94 & 0.006 & 0.04 & 1175.0 \\
\director~\cite{courant2024et} & 111.71 & 0.593 & 0.39 & 558.0 \\
\gendop\pretrainedmark~\cite{zhang2025gendop} & 339.73 & 0.188 & 0.12 & 810.0 \\
\gendop~\cite{zhang2025gendop} & 59.85 & 0.656 & 0.27 & 467.5 \\
\rowcolor{oursrow}
\textbf{\ours\ (Ours)} & \textbf{30.08} & \textbf{0.748} & \textbf{1.36} & \textbf{320.5} \\
\bottomrule
\end{tabular}
\end{table}

\paragraph{AlignScore interpretation.}
Absolute cosine values from separately trained evaluators are not directly comparable. On the same validation set, CLaTr-\dirspeed\ gives a larger matched cosine but also higher and substantially more dispersed mismatched similarities:

\begin{table}[htbp]
\centering\small
\caption{Matched and mismatched cosine statistics for the two \dirspeed\ evaluators.}
\label{tab:matched_mismatched}
\begin{tabular}{lcccc}
\toprule
Evaluator & Matched cosine\up & Mismatched cosine & R@1\up & MedR\down \\
\midrule
CLaTr-\dirspeed & 70.6 & $3.2 \pm 31.4$ & 19.7 & 23 \\
Ours-\dirspeed & 66.3 & $-0.5 \pm 22.8$ & \textbf{25.2} & \textbf{11} \\
\bottomrule
\end{tabular}
\end{table}

Accordingly, we use retrieval metrics for cross-evaluator comparison and \alnscore\ only within a fixed evaluator.

\paragraph{Absolute retrieval context.}
The validation retrieval pool contains approximately $2.5$K candidates, so random R@1 is about $0.04\%$. In addition, retrieval counts only the designated text--trajectory pair as correct even though multiple trajectories can plausibly satisfy the same motion instruction. \ours\ reaches R@1 of $3.35\%$, about $84\times$ random, and its R@5/R@10 are $12.67\%/19.79\%$, compared with $4.03\%/7.33\%$ for the strongest retrained caption-only baseline, \gendop. We therefore interpret retrieval as a discriminative alignment diagnostic rather than a claim of near-perfect instance-level matching.

\section{Ablation Studies and Extended Analysis}
\label{app:ablations}

\noindent\textbf{Design choices of \ours.}
To identify the specific drivers behind \ours's performance, we systematically ablate its key architectural and conditioning choices. As reported in Table~\ref{tab:gen_ablation}, we evaluate the necessity of our trajectory representation, multi-modal conditioning signals, and variance-guided unmasking policy.

Consistent with our core premise, replacing the \dirspeed\ representation with standard poses (\pose\ \emph{rep.}) triggers the most severe degradation across all quality, alignment, and cinematic metrics. Removing either the initial geometric anchor (\emph{w/o} $\mathbf{x}_{\mathrm{fp}}$) or the scene-level context (\emph{w/o} $\mathbf{e}_{\mathrm{logline}}$) similarly harms performance, proving that text descriptions alone are insufficient to constrain highly plausible physical paths. Finally, while random unmasking (\emph{w/o variance}) slightly elevates manifold coverage, it sacrifices crucial stability, particularly harming alignment and directorial style preservation.

\begin{table}[h]
  \centering\small
  \caption{Ablation study of \ours\ on \dataset. Underlined entries indicate specific metrics where an ablated variant marginally exceeds the full model.}
  \label{tab:gen_ablation}
  \resizebox{\linewidth}{!}{
  \begin{tabular}{lccccccccc}
    \toprule
    & \multicolumn{3}{c}{Trajectory Quality}
    & \multicolumn{3}{c}{Text--Trajectory Alignment}
    & \multicolumn{3}{c}{Movie Attributes} \\
    \cmidrule(lr){2-4}\cmidrule(lr){5-7}\cmidrule(lr){8-10}
    Setting
    & F1\up & \fcd\down & Coverage\up
    & \alnscore\up & R@1\up & MedR\down
    & Era\up & Genre\up & Director\up \\
    \midrule
    \rowcolor{oursrow}
    \ours\ (ours)
      & 0.437 & 6.77 & 0.783 & 57.79 & 3.35 & 52.0 & 47.8 & 60.7 & 48.1 \\
    \midrule
    \pose\ \emph{rep.}
      & 0.189 & 20.99 & 0.638 & 41.11 & 1.94 & 110.5 & 42.5 & 58.2 & 29.8 \\
    \emph{w/o} $\mathbf{x}_{\mathrm{fp}}$
      & 0.425 & 15.13 & 0.722 & 53.07 & \underline{3.84} & 54.0 & 40.2 & \underline{61.2} & 35.2 \\
    \emph{w/o} $\mathbf{e}_{\mathrm{logline}}$
      & 0.402 & 9.75 & 0.769 & 51.16 & 2.44 & 72.0 & 47.7 & 56.1 & 45.1 \\
    \emph{w/o variance}
      & 0.401 & 8.84 & \underline{0.784} & 53.60 & 2.79 & 58.0 & 42.1 & \underline{61.2} & 43.2 \\
    \bottomrule
  \end{tabular}}
\end{table}

\noindent\textbf{Detailed \dirspeed\ component ablation.}
To precisely isolate why \dirspeed\ yields such powerful alignment signals, we ablate its mathematical components directly within our contrastive evaluation architecture (training a new probe from scratch for each variant). In this analysis, all feature vectors refer to per-step quantities, and we omit the time index $t$ for readability. We specifically compare \emph{direction-only} features $[\mathbf{d}^{\mathrm{tr}},\mathbf{d}^{\mathrm{rot}}]$, \emph{speed-only} features $[s^{\mathrm{tr}},s^{\mathrm{rot}}]$, raw \emph{velocity} $[\Delta\mathbf{t},\boldsymbol{\omega}]$, and the full \dirspeed\ representation. 

As Table~\ref{tab:dirspd_ablation} demonstrates, normalized direction provides the vast majority of the alignment signal; relying on speed alone is insufficient for effective retrieval. However, combining both elements produces the highest R@$K$ and MedR, verifying their fundamental complementarity. Importantly, the raw \emph{velocity} baseline performs poorly. This confirms that simply taking temporal differences is not the silver bullet—the specific structural decomposition into normalized direction and log-speed is what unlocks the representation's strength.

\begin{table}[htbp]
  \centering\small
  \caption{Component ablation of \dirspeed\ for text--trajectory alignment. All variants utilize the identical contrastive architecture trained from scratch. \dirspeed\ dominates retrieval metrics, confirming the vital synergy between normalized direction and log-speed.}
  \label{tab:dirspd_ablation}
  \begin{tabular}{lccccc}
    \toprule
    Feature & R@1\up & R@5\up & R@10\up & MedR\down & \alnscore\up \\
    \midrule
    \emph{direction-only} 
      & 19.7 & 33.0 & 40.1 & 22  & 61.2 \\
    \emph{speed-only} 
      & 5.7  & 9.1  & 12.3 & 290 & 46.8 \\
    \emph{velocity} 
      & 3.8  & 8.0  & 9.3  & 377 & 48.4 \\
    \dirspeed\ (ours)
      & \textbf{25.2} & \textbf{40.7} & \textbf{49.0} & \textbf{11} & \textbf{66.3} \\
    \bottomrule
  \end{tabular}
\end{table}

\begin{table}[htbp]
  \centering\small
  \caption{Effect of text conditioning on alignment retrieval across $2{,}578$ validation clips. We train and evaluate our contrastive protocol using varying text inputs. Motion captions provide the unambiguous signal necessary for geometric alignment, whereas narrative loglines fail to act as reliable singular targets. Note: Random motion-to-text R@1 is approximately $0.04\%$.}
  \label{tab:align_gt}
  \begin{tabular}{llccccc}
    \toprule
    Rep. & Text & \alnscore\up & R@1\up & R@5\up & R@10\up & MedR\down \\
    \midrule
    \multirow{3}{*}{\dirspeed}
      & motion           & \textbf{66.3} & \textbf{25.2} & \textbf{40.7} & \textbf{49.0} & \textbf{11} \\
      & motion + logline & 50.6          & 4.6           & 14.6          & 23.7          & 41 \\
      & logline          & 31.1          & 1.0           & 2.9           & 4.8           & 227 \\
    \midrule
    \multirow{3}{*}{\pose}
      & motion           & 38.7          & 8.7           & 15.9          & 21.5          & 66 \\
      & motion + logline & 33.3          & 2.3           & 7.7           & 12.3          & 102 \\
      & logline          & 26.2          & 0.7           & 1.9           & 3.0           & 428 \\
    \bottomrule
  \end{tabular}
\end{table}

\subsection{Representation Robustness and Matched Controls}
\label{app:repr_robustness}

\paragraph{Coordinate and scale controls.}
Our default \pose\ representation centers translation at the first frame but retains world-frame rotation, whereas \dirspeed\ uses world-frame translation increments and camera-local rotation increments. To test whether the advantage comes merely from removing coordinate gauge or trajectory scale, we additionally canonicalize \pose\ to the first camera frame and then apply the trajectory-scale normalization used by \gendop, dividing translations by $\max_t\lVert\mathbf{t}_t-\mathbf{t}_1\rVert_2$.

\begin{table}[htbp]
\centering\small
\caption{Matched coordinate and scale controls for \pose.}
\label{tab:pose_controls}
\begin{tabular}{lccc}
\toprule
Representation & \alnscore\up & R@1\up & MedR\down \\
\midrule
\emph{First-frame-canonical} \pose & 46.4 & 15.2 & 51 \\
+ \emph{Scale normalization} \pose & 49.0 & 18.9 & 40 \\
\dirspeed & \textbf{66.3} & \textbf{25.2} & \textbf{11} \\
\bottomrule
\end{tabular}
\end{table}

Canonicalization provides only a limited gain and scale normalization a moderate additional improvement, while a substantial gap to \dirspeed\ remains. Thus, the representation advantage is not explained solely by coordinate or scale conventions.

\paragraph{Speed parameterization.}
We next keep the normalized direction features fixed and vary only the speed transformation. All variants use the same split, evaluator architecture, and training protocol.

\begin{table}[htbp]
\centering\small
\caption{Ablation of the speed parameterization with direction features held fixed.}
\label{tab:speed_transform}
\begin{tabular}{lccc}
\toprule
Magnitude parameterization & \alnscore\up & R@1\up & MedR\down \\
\midrule
Linear speed & 63.6 & 21.5 & 14 \\
Global z-score & 64.7 & 24.4 & 11 \\
$\log(1+\mathrm{speed})$ & 64.5 & 22.4 & 14 \\
$\log(\mathrm{speed}+\varepsilon)$ & \textbf{66.3} & \textbf{25.2} & \textbf{11} \\
\bottomrule
\end{tabular}
\end{table}

The relatively narrow range across these variants confirms that normalized direction carries most of the alignment signal. The default logarithmic parameterization gives the best overall result and compresses the long-tailed distribution of camera-motion magnitudes.

\paragraph{Reconstruction and long-range fidelity.}
Given the initial camera-to-world pose $(R_1,\mathbf{t}_1)$, a \dirspeed\
sequence is converted back to poses by normalizing the predicted directions
and recovering the increments
\[
\Delta\mathbf{t}_t
=
\bar{\mathbf d}^{\mathrm{tr}}_t\bigl(\exp(s^{\mathrm{tr}}_t)-\varepsilon\bigr),
\qquad
\Delta R_t
=
\operatorname{Exp}\!\left(
\bar{\mathbf d}^{\mathrm{rot}}_t\bigl(\exp(s^{\mathrm{rot}}_t)-\varepsilon\bigr)
\right),
\]
where $\bar{\mathbf d}$ denotes the normalized predicted direction and
$\operatorname{Exp}=\exp_{SO(3)}$. We then recursively apply
\[
\mathbf t_t=\mathbf t_{t-1}+\Delta\mathbf t_t,
\qquad
R_t=R_{t-1}\Delta R_t.
\]
A pose $\rightarrow$ \dirspeed $\rightarrow$ pose round trip on validation
trajectories longer than $200$ frames yields a median endpoint error of
$3.84\times10^{-9}$ in translation and $0.130^\circ$ in rotation. We also
split generated outputs into short ($59$--$95$ frames) and long
($\geq 213$ frames) sequences:

\begin{table}[htbp]
\centering\small
\caption{Generation quality on short and long trajectories.}
\label{tab:long_range}
\begin{tabular}{lcccccc}
\toprule
Representation & Short F1\up & Long F1\up & Short \fcd\down & Long \fcd\down & Short Cov.\up & Long Cov.\up \\
\midrule
\pose & 0.240 & 0.187 & \textbf{10.14} & 33.08 & 0.794 & 0.635 \\
\dirspeed & \textbf{0.411} & \textbf{0.419} & 10.72 & \textbf{17.72} & \textbf{0.820} & \textbf{0.786} \\
\bottomrule
\end{tabular}
\end{table}

\dirspeed\ remains stable as the sequence horizon grows, while \pose\ degrades substantially on the long subset.

\paragraph{Joint pose--motion representation.}
Absolute pose can still provide useful complementary information for tasks where global spatial context matters. Concatenating a canonicalized \pose\ stream with \dirspeed\ gives a small additional gain, while \dirspeed\ alone captures most of the improvement with less than half the input dimensionality.

\begin{table}[htbp]
\centering\small
\caption{Joint use of pose and motion features.}
\label{tab:joint_rep}
\begin{tabular}{lcccc}
\toprule
Representation & Input dim. & \alnscore\up & R@1\up & MedR\down \\
\midrule
\pose & 9 & 46.6 & 17.8 & 51 \\
\dirspeed & 8 & 66.3 & 25.2 & 11 \\
\pose+\dirspeed & 17 & \textbf{66.9} & \textbf{27.0} & \textbf{9} \\
\bottomrule
\end{tabular}
\end{table}

\paragraph{Data and model scaling.}
Finally, we compare how the two representations use additional data and model capacity. \dirspeed\ benefits more consistently from larger training sets; with only $50\%$ of the data it already exceeds \pose\ trained on the full set. Increasing encoder size beyond the base model yields little further improvement for either representation.

\begin{table}[htbp]
\centering\small
\caption{Data scaling for \pose\ and \dirspeed.}
\label{tab:data_scaling}
\begin{tabular}{ccc}
\toprule
Training data & \pose\ R@1\up & \dirspeed\ R@1\up \\
\midrule
10\% & 11.3 & 15.1 \\
25\% & 11.0 & 15.9 \\
50\% & 12.2 & 20.2 \\
100\% & 17.8 & 25.2 \\
\bottomrule
\end{tabular}
\end{table}
\begin{table}[htbp]
\centering\small
\caption{Trajectory-encoder scaling.}
\label{tab:model_scaling}
\begin{tabular}{lccc}
\toprule
Model & Trainable params & \pose\ R@1\up & \dirspeed\ R@1\up \\
\midrule
Small & 0.53M & 16.3 & 24.2 \\
Base & 3.49M & 17.8 & 25.2 \\
Large & 26.14M & 17.1 & 25.0 \\
\bottomrule
\end{tabular}
\end{table}

\section{Experimental Settings}
\label{app:experimental_settings}

This section provides the comprehensive hyperparameters and architectural details for both our contrastive alignment evaluator and the \ours\ generative model. All experiments are conducted on a single NVIDIA A100 GPU. Due to the differences in architectural complexity, the training costs vary significantly: our lightweight contrastive evaluator converges rapidly, typically within $30$ epochs (requiring only ${\sim}10$ minutes of wall-clock time), whereas training the \ours\ generative model typically requires $150$ to $200$ epochs, totaling approximately $12$ hours.

\subsection{Contrastive Alignment Evaluator Configurations}

The contrastive evaluator is trained using a symmetric InfoNCE loss with a fixed temperature of $0.1$. To prevent penalizing semantically identical captions, we apply false-negative filtering by masking off-diagonal pairs whose text-to-text cosine similarity exceeds $0.99$. Unlike prior evaluators (e.g., CLaTr), our protocol strictly avoids any trajectory reconstruction, KL divergence, or decoder overhead. Detailed hyperparameters are listed in Table~\ref{tab:evaluator_config}.

\subsection{\ours\ Generative Model Configurations}

\ours\ is trained to predict denoised trajectory tokens conditioned on multimodal text inputs and initial geometric anchors. The text conditioning uses a combined approach, encoding the motion caption and logline separately. The model comprises approximately $28.4$M trainable parameters (excluding the frozen text encoder and EMA copies). Detailed hyperparameters are summarized in Table~\ref{tab:generator_config}.

\begin{table}[h]
  \centering\small
  \caption{Architecture and training hyperparameters for the Contrastive Alignment Evaluator.}
  \label{tab:evaluator_config}
  \begin{tabular}{ll}
    \toprule
    \textbf{Hyperparameter} & \textbf{Value} \\
    \midrule
    \multicolumn{2}{l}{\emph{Trajectory Encoder (Trainable, ${\sim}3.6$M params)}} \\
    Input dimension & 8 (\dirspeed) or 9 (\pose) \\
    Linear projection & $256$ \\
    Positional encoding & Sinusoidal (max length $5000$) \\
    Transformer layers & $4$ \\
    Transformer dimensions & $d_{\mathrm{model}}=256$, $4$ heads, $d_{\mathrm{ff}}=1024$ \\
    Transformer configuration & Pre-LayerNorm, Dropout $0.1$, GELU \\
    Output pooling & Learnable [CLS] token prepended \\
    Output head & LayerNorm $\rightarrow$ Linear $\rightarrow$ GELU $\rightarrow$ Linear ($256$ dims) \\
    \midrule
    \multicolumn{2}{l}{\emph{Text Encoder (Frozen Backbone, ${\sim}197$K trainable params)}} \\
    Backbone & CLIP ViT-L/14 (frozen) \\
    Tokens & Max $77$, using pooled [EOS] embedding ($768$ dims) \\
    Projection head & Linear ($768 \rightarrow 256$) \\
    \midrule
    \multicolumn{2}{l}{\emph{Training and Optimization}} \\
    Optimizer & AdamW ($\beta_1=0.9$, $\beta_2=0.999$, weight decay $= 10^{-4}$) \\
    Learning rate & $2 \times 10^{-4}$ with Cosine Annealing \\
    Batch size & $64$ \\
    Gradient clipping & Max norm $1.0$ \\
    Epochs & $150$ (Early stopping patience $= 10$) \\
    \bottomrule
  \end{tabular}
\end{table}

\begin{table}[htbp]
  \centering\small
  \caption{Architecture, sampling, and training hyperparameters for \ours.}
  \label{tab:generator_config}
  \begin{tabular}{ll}
    \toprule
    \textbf{Hyperparameter} & \textbf{Value} \\
    \midrule
    \multicolumn{2}{l}{\emph{Conditioning and Architecture}} \\
    Text encoder & CLIP ViT-B/32 (frozen, $512$ dims) \\
    Logline embedding dim & $64$ \\
    Sequencer & TransformerAdaLN ($1$ block) \\
    Sequencer dimensions & $d_{\mathrm{model}}=512$, $8$ heads, $d_{\mathrm{ff}}=4096$, Dropout $0.2$ \\
    Conditioning fusion & Text ($512$) + First Pose ($64$) + Logline ($64$) = $640$ dims $\rightarrow$ $512$ dims \\
    Diffuser & SimpleMLPAdaLN ($3$ ResBlocks, $d_{\mathrm{model}}=1024$) \\
    \midrule
    \multicolumn{2}{l}{\emph{Masking and Sampling}} \\
    Masking ratio range & $[0.5, 1.0]$ uniformly sampled during training \\
    Unmasking order & Variance-guided \\
    Autoregressive (AR) steps & $18$ \\
    DDPM scheduling & Cosine ($\mathrm{s}=0.008$), $T=100$ \\
    DDPM sampling steps & $50$ \\
    Classifier-Free Guidance & Drop prob $= 0.1$ (train), CFG scale $= 3.5$ (inference) \\
    \midrule
    \multicolumn{2}{l}{\emph{Training and Optimization}} \\
    Optimizer & AdamW ($\beta_1=0.9$, $\beta_2=0.99$, weight decay $= 10^{-5}$) \\
    Learning rate & $3 \times 10^{-4}$ (Linear warmup for $2000$ steps, then constant) \\
    Batch size & $128$ \\
    Batch multiplier & $5$ (noise levels per masked token) \\
    Gradient clipping & Max norm $1.0$ \\
    Epochs & $500$ (EMA decay $= 0.999$) \\
    \bottomrule
  \end{tabular}
\end{table}

\section{The Complementary Roles of Motion Captions and Loglines}
\label{app:text_choice_alignment}

Throughout the main text, we consistently train and evaluate alignment using motion captions. This section provides the empirical justification for isolating this specific textual signal, while also examining the role of loglines during generation.

\noindent\textbf{Alignment requires strict geometric correspondence.}
Given that motion captions explicitly describe mechanical camera behavior, they form a tightly coupled correspondence with the physical trajectory. Loglines, in contrast, describe contextual scene elements. Table~\ref{tab:align_gt} compares our contrastive protocol trained under three distinct text inputs: isolated motion captions, concatenated motion and loglines, and isolated loglines. 

The empirical results perfectly align with our intuition. Motion captions yield by far the strongest text--trajectory retrieval across both \dirspeed\ and \pose\ architectures. Diluting the motion caption with loglines degrades retrieval performance, and relying on loglines alone leads to a near-collapse in recall capability. This demonstrates that contextual scene descriptions are too underdetermined to serve as a direct geometric alignment target. 

\noindent\textbf{Generation benefits from narrative context.}
However, this lack of strict one-to-one geometric correlation does not render loglines useless. As established by our generation metrics (Table~\ref{tab:gen_ablation}), using loglines as an auxiliary input affects trajectory realism and the preservation of movie-attribute correlations beyond what is captured by the motion caption alone.

\begin{figure}[htbp]
  \centering
  \includegraphics[width=\linewidth]{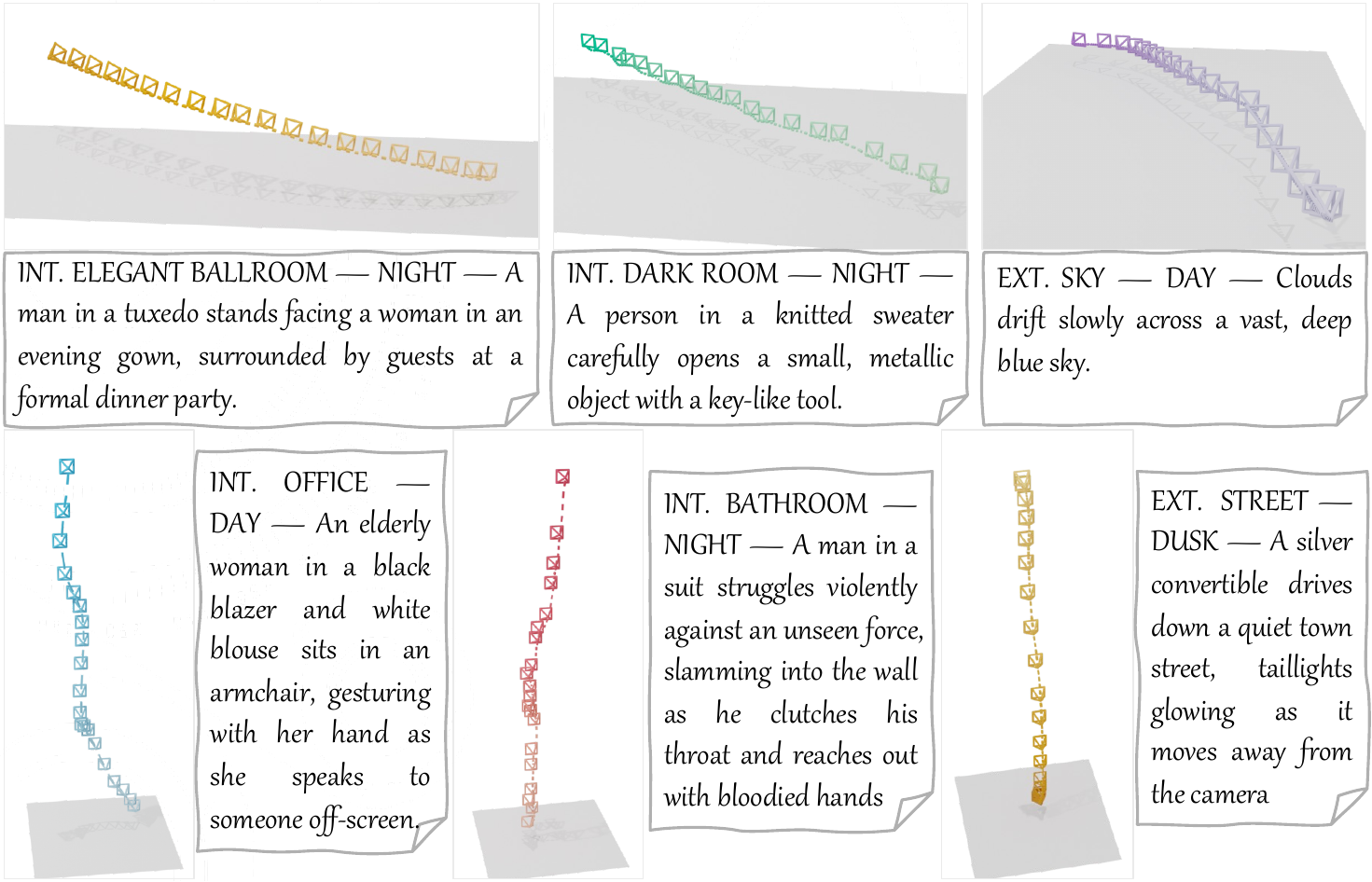}
  \caption{Impact of logline conditioning on trajectory generation. We generate camera paths using the same motion caption (Top row: ``truck left'', Bottom row: ``pedestal up'') but varying loglines. While the fundamental camera movement firmly aligns with the motion caption, the logline modulates the trajectory's scale, pace, and subtle dynamics to suit the specific narrative context.}
  \label{fig:traj_logline}
\end{figure}

To qualitatively illustrate this effect, Figure~\ref{fig:traj_logline} visualizes trajectories generated by \ours\ conditioned on a fixed motion caption but paired with different scene loglines. The top row shows variations under a ``truck left'' instruction, while the bottom row uses ``pedestal up.'' Although the base instruction remains fixed, the logline can modulate the trajectory's scale, pace, and subtle dynamics. These examples illustrate scene-conditioned variation; they do not constitute explicit era-, genre-, or director-style control.

\noindent\textbf{Robustness to conflicting logline cues.}
The annotation prompt excludes camera-motion language to keep the two text conditions semantically distinct, but inference does not require a sanitized logline. To test robustness, we select $128$ validation samples with explicit directional motion captions and inject the \emph{opposite} direction into the corresponding logline while holding the motion caption, first pose, and initial sampling noise fixed. Similarity to the original motion caption does not decrease ($0.574$ with the clean logline versus $0.592$ with the contradictory logline). Moreover, in $90/128$ cases ($70.3\%$), the generated trajectory remains closer to the dedicated motion-caption direction than to the conflicting logline direction. This stress test indicates that the motion caption remains the dominant control signal even when the auxiliary scene description contains an explicit conflict.

\section{Limitations and Broader Impacts}
\label{app:limitations_impacts}

\paragraph{Limitations.}
While \ours\ establishes a new paradigm for motion-centric camera trajectory generation, it has several limitations that provide promising avenues for future work. 
First, our exploration of cinematic attributes is constrained by data availability. The subset of \dataset\ with explicitly linked real-movie metadata (e.g., era, genre, director) is relatively small (${\sim}3.2$K clips) and severely long-tailed, making reliable supervised style control difficult. We therefore use these attributes to evaluate preservation of latent cinematic correlations rather than to train an era-, genre-, or director-controlled generator. Scaling and balancing the metadata-linked subset is an important direction for explicit controllable cinematic styling.
Second, like all data-driven motion models, our framework relies on the accuracy of the underlying Structure-from-Motion (SfM) or SLAM pipeline~\cite{vipe} used to extract camera poses from raw video. Our filtering, smoothing, and caption-tag stabilization remove severe failures and isolated jitter before training, but some residual estimation noise can remain.
Finally, while our VLM-generated loglines provide valuable scene context, they are inherently synthetic. Potential hallucinations or missed subtle narrative cues from the VLM could introduce noise into the text-conditioning space.

\paragraph{Broader Impacts.}
Our research holds significant positive potential for virtual production, 3D animation, and independent filmmaking. By allowing creators to synthesize complex, realistic camera movements using natural language and scene descriptions, \ours\ can democratize pre-visualization and reduce the steep learning curve associated with professional 3D camera rigging. 
Conversely, as with any generative video technology, improved camera trajectory synthesis could theoretically be dual-used to enhance the realism of synthetic media or deepfakes, making them harder to distinguish from real footage. However, we note that our model specifically outputs abstract physical parameters (camera trajectories) rather than rendering raw visual pixels. To generate misleading video content, our framework would need to be coupled with a high-fidelity rendering engine or video generation model. We advocate for the continued development of robust watermarking and synthetic-media detection protocols to mitigate these downstream risks.

\end{document}